\documentclass[10pt,journal,final,twocolumn]{IEEEtran}

\usepackage{graphicx,times,amsmath,amssymb,cite,cases,mathrsfs,booktabs,multirow,psfrag,subfigure,url,stfloats,threeparttable,epstopdf,pdfpages}
\usepackage{algorithm,algorithmic,amsthm,ulem,latexsym,bm}
\usepackage{dsfont}
\usepackage[colorlinks,linkcolor=red,anchorcolor=blue,citecolor=red]{hyperref}
\newcommand{\tabincell}[2]{\begin{tabular}{@{}#1@{}}#2\end{tabular}}

\begin{document}

\title{Bures Joint Distribution Alignment with Dynamic Margin for Unsupervised Domain Adaptation}
\author{Yong-Hui Liu, Chuan-Xian Ren, Xiao-Lin Xu, Ke-Kun Huang 
\thanks{Y.H. Liu and C.X. Ren are with the School of Mathematics, Sun Yat-Sen University, Guangzhou, 510275, China. C.X. Ren is the corresponding author (email: rchuanx@mail.sysu.edu.cn). X.L. Xu is with the School of Statistics and Mathematics, Guangdong University of Finance and Economics, Guangzhou 510320, China. K.K. Huang is with the School of Mathematics, JiaYing University, Meizhou, 514015, China. }}

\date{}
\IEEEcompsoctitleabstractindextext{%

\begin{abstract}
Unsupervised domain adaptation (UDA) is one of the prominent tasks of transfer learning, and it provides an effective approach to mitigate the distribution shift between the labeled source domain and the unlabeled target domain. Prior works mainly focus on aligning the marginal distributions or the estimated class-conditional distributions. However, the joint dependency among the feature and the label is crucial for the adaptation task and is not fully exploited. To address this problem, we propose the Bures Joint Distribution Alignment (BJDA) algorithm which directly models the joint distribution shift based on the optimal transport theory in the infinite-dimensional kernel spaces. Specifically, we propose a novel alignment loss term that minimizes the kernel Bures-Wasserstein distance between the joint distributions. Technically, BJDA can effectively capture the nonlinear structures underlying the data. In addition, we introduce a dynamic margin in contrastive learning phase to flexibly characterize the class separability and improve the discriminative ability of representations. It also avoids the cross-validation procedure to determine the margin parameter in traditional triplet loss based methods. Extensive experiments show that BJDA is very effective for the UDA tasks, as it outperforms state-of-the-art algorithms in most experimental settings. In particular, BJDA improves the average accuracy of UDA tasks by 2.8\% on Adaptiope, 1.4\% on Office-Caltech10, and 1.1\% on ImageCLEF-DA.
\end{abstract}

\begin{IEEEkeywords}
Unsupervised domain adaptation, Joint distribution, Optimal transportation, Contrastive learning, Dynamic margin.
\end{IEEEkeywords}}

\maketitle \IEEEdisplaynotcompsoctitleabstractindextext \IEEEpeerreviewmaketitle

\section{Introduction}\label{section1}

\begin{figure}[tb]
\centering\includegraphics[scale=0.7, trim=2 2 4 6,clip]{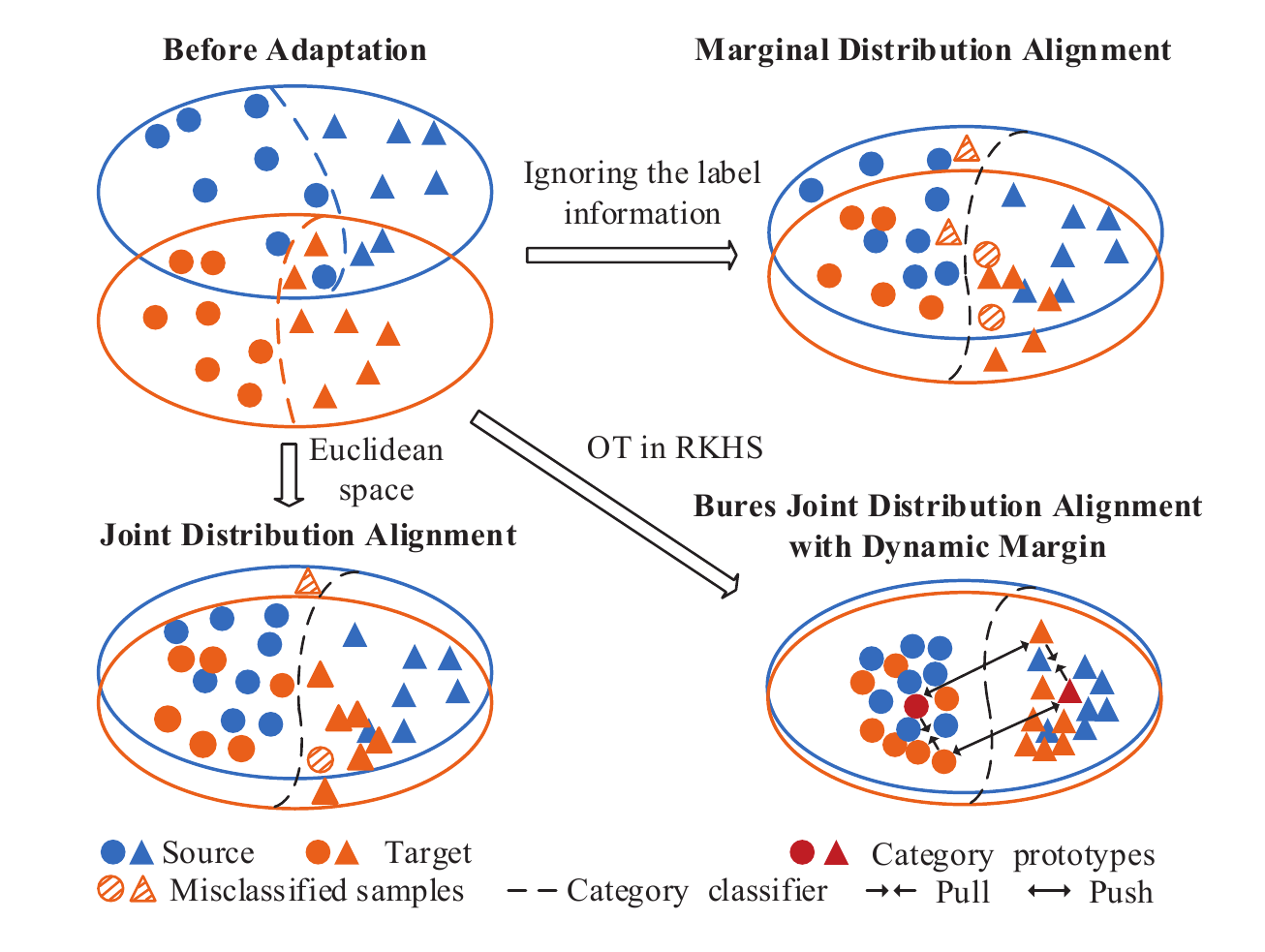}
\caption{Comparisons of existing methods and the proposed BJDA method. Dots and triangles denote different categories. Top left: there exists domain shift between the source and target domains, and different domains specializes in different category classifiers (i.e., the dashed lines with different colors). Top right: some UDA methods perform \textit{marginal distribution} alignment while ignoring the label information, which cannot address the domain shift well. Bottom left: some approaches perform joint distribution alignment in the Euclidean space, but the performance is limited. Bottom right: our BJDA can effectively exploit the nonlinear structures underlying the data in RKHS, and the proposed dynamic margin based contrastive learning phase prompts the samples to get close to the category  prototype of the same label and keep away from the category prototypes of different labels. Better viewed in color.}\label{wjda1}
\end{figure}

\IEEEPARstart{T}{he} generalization ability across datasets is a long-standing goal of machine learning models. While deep neural networks have demonstrated the powerful generalization ability on a number of tasks, these remarkable gains largely rely on the availability of sufficient labeled training data. Annotating large amounts of data is unfortunately difficult or expensive in real-world applications. One natural approach \cite{Lu2016tip} is to transfer the learned knowledge from one dataset to another new dataset. However, various datasets are collected under different conditions. Thus, the distribution shift across datasets often leads to significant performance degradation, as shown in the top left of Fig. \ref{wjda1}. Unsupervised domain adaptation (UDA) \cite{pr1,pr4,CKB2021} specifically deals with this situation, and learns a model for the unlabeled test data (i.e., the target domain) by transferring knowledge from the labeled training data (i.e., the source domain).

UDA has been investigated for many fields including object recognition \cite{pr3,Ren2020TCYB1,TWMDApr}, medical image diagnosis \cite{TanMK2020,Gengxin2021}, and natural language processing \cite{JDOT2017,PR2021doc1}. Let $\Omega$ be the input space and $\mathcal{C}$ the label space, $X_{s},X_{t} \subset \Omega$ denote the source and target samples, $Y_{s},Y_{t} \subset \mathcal{C}$ denote the corresponding labels, where $Y_{t}$ is unavailable during training. As shown in the top right of Fig. \ref{wjda1}, many previous works focus on reducing the difference between the marginal feature distributions of the two domains, i.e., $P(X_s)$ and $P(X_t)$, as small as possible. In some cases \cite{coral2016AAAI,Ren2021cov}, statistical moments such as expectation and covariance matrix are used to model the difference between the feature distributions of the two domains. Then, this idea has been extended to use some statistical distance to measure the discrepancy of feature distributions from various domains, such as the maximum mean discrepancy (MMD) \cite{DDC2014,dubo2020tip2,ZhLei2021,pr2}, or the Wasserstein distance \cite{Li2020etd}. Some works propose the domain-adversarial strategy \cite{Yaroslav2016JMLR,TanMK2019,dubo2020Learning} to align the feature distributions with a domain discriminator. While the above approaches try to learn domain-invariant features between source and target domains, they rarely take the label information into consideration during the adaptation process. Another line of research \cite{JDA2013,Yang2020tip} involves the conditional distributions of labels, i.e., $P(Y_{s}|X_{s})$ and $P(Y_{t}|X_{t})$, which attempts to minimize the difference in both the marginal feature distributions and the conditional distributions between the source and the target domains. Since $P(Y_{s}|X_{s})$ and $P(Y_{t}|X_{t})$ are very difficult to estimate, they usually use the class-conditional distributions $Q(X_{s}|Y_{s})$ and $Q(X_{t}|Y_{t})$  instead, which limits their effectiveness.

Motivated by the fact that the features and labels are jointly drawn from certain high-dimensional probability distributions, some methods have been designed to directly address the difference between the joint distributions of the two domains, as shown in the bottom left of Fig. \ref{wjda1}. Courty et al. \cite{JDOT2017} assume that there exists an optimal transport transformation between the joint distributions across different domains, i.e., $P(X_{s}, Y_{s})$ and $P(X_{t},Y_{t})$, and propose joint distribution optimal transport (JDOT) to align the joint distributions. Damodaran et al. \cite{DeJDOT2018} incorporate JDOT into deep neural networks and propose deep joint distribution optimal transport (DeepJDOT) for UDA. Although they have shown promising performance in UDA, there are still some limitations. On the one hand, they rely on the alternate optimization strategy for the optimal transport plan and the classifier, which is unstable and hard to train. On the other hand, these works typically reduce the distribution discrepancy in the Euclidean space, which may ignore the nonlinear structures underlying the data.

To address these issues mentioned above, we introduce a novel Bures joint Distribution Alignment (BJDA) approach based on the optimal transport (OT) theory (see the bottom right of Fig. \ref{wjda1}). The key idea is to model the joint distribution shift in the reproducing kernel Hilbert space (RKHS). In addition, we introduce a dynamic margin based contrastive learning phase to improve the category discriminative ability of representations. Importantly, we use an equivalent and computable formulation \cite{OTrkhs} of the kernel Bures-Wasserstein distance for the discrepancy between the joint distributions by generalizing the original optimal transport problem in Euclidean space to kernel space. Thus, our method can capture the nonlinear structures underlying the data, and it provides an efficient and end-to-end trainable approach for reducing the joint distribution shift between the source domain and the target domain.

The main contributions of this paper are summarized as follows:
\begin{enumerate}
\item{A Bures joint distribution alignment phase is proposed, which minimizes the discrepancy of joint distributions between the source and target domains in the infinite-dimensional kernel space.}
\item{A dynamic margin based contrastive learning phase is proposed, which enhances the discriminative ability of the features and avoids cross-validation to determine the margin parameters. }
\item{The proposed BJDA method is evaluated on several object classification tasks, and it achieves state-of-the-art classification performance on four publicly available datasets.}
\end{enumerate}

The remainder of the paper is organized as follows. In Section \ref{section2}, some closely related works are reviewed. Algorithmic details of our BJDA method are presented in Section \ref{section3}. Section \ref{section4} provides extensive experimental results for validating the superiority of BJDA. Finally, we conclude the paper with a brief discussion in Section \ref{section5}.

\section{Related Works}\label{section2}

In this section, we briefly review some recent works about UDA by considering the marginal distribution and the joint distribution, respectively. While many previous works exist, we focus primarily on deep learning-based methods. In light of our method is based on the optimal transport theory, we also review some progress about OT.

\subsection{Unsupervised Domain Adaptation with Marginal Distribution Shift}

Most recent UDA methods are motivated by the seminal theoretical work of Ben-David et al. \cite{Ben2006NIPS}, which proposes the $\mathscr{A}$-distance to estimate the discrepancy of feature marginal distributions between source and target domains. Then Blitzer et al. \cite{Blitzer2007NIPS} develop the $\mathscr{H}$-distance to deduce a uniform learning bound, which minimizes the convex combination of empirical risks among various domains. A thorough theoretical investigation for UDA is presented in \cite{Ben2010MacLen} in terms of the $VC$ dimension. In light of these foundation works, minimizing the domain distribution discrepancy has been used extensively for UDA. The covariance matrix is used in several works \cite{coral2016AAAI,Ren2021cov} to model the difference between the feature distributions of various domains. Ren et al. \cite{Ren2020TCYB1} exploit a low-rank representation to reduce the marginal distribution shift and stress the group compactness of feature representations. Li et al. \cite{AdaBN2018pr} design an adaptive batch normalization operation to align the feature distributions of different domains. Luo et al. \cite{YouWeiPami1} use a sequence of latent manifolds to describe the diverse domains and employ the manifold metrics to model the domain discrepancy. While methods discussed above are mainly based on distribution discrepancy, there have been several works to address the problem of UDA by applying adversarial learning. Ganin et al. \cite{Yaroslav2016JMLR} introduce domain adversarial neural network (DANN) to align the feature distributions with a domain discriminator via the adversarial mechanism. Tzeng et al. \cite{Tzeng2017CVPRadda} use separate feature extraction networks for diverse domains and train the target domain adversarially until it is adapted to the source domain. Satio et al. \cite{MCD2018} propose maximum classifier discrepancy (MCD), which utilizes two category classifiers to simulate a domain discriminator and train the feature generator in an adversarial fashion. These approaches focus on learning a domain-invariant feature representation among the source and target domains, the available label information from the source domain is not fully exploited during the adaptation process.

\subsection{Unsupervised Domain Adaptation with Joint Distribution Shift}

Domain shift usually exists both marginal distributions and conditional distributions. Thus, it is more important to investigate the domain difference among the joint distributions. To account for the label information, Li et al.~\cite{Atm2021} propose a novel domain distance, Maximum Density Divergence (MDD), which measures both the inter-domain divergence and the intra-class density. Li et al.~\cite{DivAg2021} also propose Adaptation by Adversarial Attacks (AAA), which addresses the UDA task from the view of the adversarial attack, and generalizes the UDA task to the scenario in which the source data or the target data is absent during training. Then, Faster Domain Adaptation (FDA)~\cite{FDA2021} is proposed, which adaptively selects the number of neural network layers for various samples and significantly speeds up the adaptation process. Li et al.~\cite{HetDA2019} further propose Heterogeneous Domain Adaptation (HDA), which uses dictionary learning to handle heterogeneous feature spaces with different dimensions, and minimizes the marginal distribution gaps across different domains. Zhang et al.~\cite{JointDA2021} use a domain discriminator to align the marginal distribution with adversarial learning, and develop a class-wise divergence to address the gap between conditional distributions. Du et al.~\cite{Unified2021} align the marginal distribution with the domain discriminator, it also utilizes the disagreement of two category classifiers to measure the discrepancy of conditional distributions. Yu et al.~\cite{Dynamic2019} use two domain discriminators, one for the marginal distributions, and the other for the conditional distributions. Wang et al.~\cite{Dynamic2020} propose Dynamic Distribution Adaptation (DDA) to address the different effects of marginal and conditional distributions in the adaptation process. Wang et al.~\cite{Jindong2018meda} train a domain-invariant classifier in Grassmann manifold and minimizes the structural risk. Luo et al.~\cite{Look2019} propose a category-level adversarial network to enforce local semantic consistency. Then, Luo et al.~\cite{CLaa2021} introduce a novel Significance-aware Information Bottleneck (SIB) to improve the former-mentioned category-level adaptation. Luo et al.~\cite{SA2019} further propose a novel Significance Aware Layer, which uses the channel-wise significance of each semantic feature to balance the information from different categories. Although these works take the label information into consideration, they align the marginal distributions and the conditional distributions separately.

Long et al. \cite{JDA2013} propose Joint Distribution Adaptation (JDA) by aligning both the marginal feature distributions and the conditional distributions in a dimensionality reduction procedure. Chen et al. \cite{Yang2020tip} exploit nonlinear projections to match the projected distributions. Long et al. \cite{Long2017JAN} propose the Joint Maximum Mean Discrepancy (JMMD) to measure the discrepancy of the joint distributions of feature representations from different layers in the neural network between diverse domains. Li et al. \cite{UDA2018lishuang} propose to match the marginal and conditional distributions via the maximum mean discrepancy (MMD) distance. Long et al. \cite{cdan1} develop a multilinear map to model the interactions across the feature and label predictions. Zhang et al. \cite{TanMK2019} propose a symmetric framework to perform domain confusion both on label-level and feature-level. Gu et al. \cite{Rsda2020} take the label information into consideration and propose to correct the pseudo-labels based on the feature distance. Ni et al. \cite{Joint2021Z} propose to match both the marginal distributions and conditional distributions by training a domain-invariant kernel. Courty et al. \cite{JDOT2017} investigate an optimal transport transformation to align the joint distributions between different domains and propose joint distribution optimal transport (JDOT). Damodaran et al. \cite{DeJDOT2018} incorporate JDOT with deep neural networks and propose deep joint distribution optimal transport (DeepJDOT). Although these works have shown promising performance in UDA, some of them reduce the distribution discrepancy in the Euclidean space, which may ignore the nonlinear structures underlying the data, and the others usually train the deep neural networks and the optimal transport task alternately, which is unstable and hard to train.

\subsection{UDA with Optimal Transport}

Optimal transportation (OT) \cite{VilaniOT} has been an important topic in both the computer vision and machine learning communities. OT investigates how to effectively transport a probability distribution to another one, and the total cost during the transportation is related to a distance between diverse probability distributions. Assume that we are given two probability measures $\mu$ and $\nu$, defined respectively on some measure spaces $\mathcal{Z}_{1}$ and $\mathcal{Z}_{2}$.  The cost function $D(\mathbf{z}_{1}; \mathbf{z}_{2})$ defined on $\mathcal{Z}_{1} \times \mathcal{Z}_{2}$ depicts how much it costs to transport a unit of probability mass from the point $\mathbf{z}_{1}$ to the point $\mathbf{z}_{2}$, where $\mathbf{z}_{1} \in \mathcal{Z}_{1}$ and $\mathbf{z}_{2} \in \mathcal{Z}_{2}$. The transport coupling $\gamma (\mathbf{z}_{1}; \mathbf{z}_{2})$ is a joint probability measure which can be interpreted as the amount of probability mass transported from the point $\mathbf{z}_{1}$ to the point $\mathbf{z}_{2}$. The optimal transport cost between the two measures $\mu$ and $\nu$ defines the Wasserstein distance \cite{VilaniOT}, which is denoted as
\begin{equation}
\label{eq1w}
d_{W}(\mu,\nu)=\left[\inf_{\gamma \in \Pi(\mu,\nu) } \int_{\mathcal{Z}_{1} \times \mathcal{Z}_{2}} D(\mathbf{z}_{1}; \mathbf{z}_{2}) d \gamma(\mathbf{z}_{1}; \mathbf{z}_{2})\right]^{1/2}.
\end{equation}

Courty et al. \cite{OTuda1} propose a regularized optimal transportation algorithm to perform the alignment of the feature representations among the source and target domains. Yang et al.~\cite{qian2020JOT} present a novel robust regression scheme by integrating OT with convex regularization terms. Li et al. \cite{Li2020etd} use an attention-aware transport distance to directly minimize the domain discrepancy. Usually, it is complicated to compute the Wasserstein distance as it requires for solving a linear programming task iteratively. Thus, it is not straightforward to incorporate the Wasserstein distance with deep neural networks. To address this issue, Bhatia et al. \cite{Bures2019} extend the Wasserstein distance to the Bures-Wasserstein distance, which is easy and computable. Let $S^+(n)$ be the set of all PSD matrices of size $n \times n$. The Bures-Wasserstein distance \cite{Bures2019} between two probability measures $\mu$ and $\nu$ is defined as
\begin{equation}
\label{eq2bw}
d_{BW}(\mu, \nu)=\left[\operatorname{tr}(\mathbf{\Sigma}_{1})+\operatorname{tr}(\mathbf{\Sigma}_{2})-2\operatorname{tr}(\mathbf{\Sigma}_{1}^{1/2}\mathbf{\Sigma}_{2}\mathbf{\Sigma}_{1}^{1/2})^{1/2}\right]^{1/2},
\end{equation}
where $\mathbf{\Sigma}_{1}$, $\mathbf{\Sigma}_{2} \in S^+(n)$ are the covariance matrices of $\mu$ and $\nu$, respectively. The original OT problem is discussed in the Euclidean space. Recently, several works \cite{OTrkhs,CKB2021} generalize the original optimal transport method to infinite-dimensional feature spaces. An equivalent and computable closed-form formulation \cite{OTrkhs} of the Kantorovich optimal transport problem is provided via the kernel methods, which provides a probable solution for measuring the discrepancy among diverse distributions. The theorem in \cite{OTrkhs} states that if $\mu$ and $\nu$ are two Gaussian measures with the same expectation, the Wasserstein distance is just the corresponding Bures-Wasserstein distance between their covariance matrices on the manifold of Positive Semi-Definite (PSD) matrices.

\section{Bures joint Distribution Alignment}\label{section3}

In this section, we describe the BJDA algorithm in details. We first present basic formulation for the UDA problem. Then we introduce the Bures joint distribution alignment phase (loss term) and the dynamic margin based contrastive learning phase.

\subsection{Problem Formulation}

Without loss of generality, we consider a $C$-class problem as an example. Let $\Omega$ be the input space and $\mathcal{C}$ the label space, where $\mathcal{C} = \{1,\cdot\cdot\cdot,C \}$. $X_{s},X_{t} \subset \Omega$ denote the source and target samples, $Y_{s},Y_{t} \subset \mathcal{C}$ denote the label of source and target samples, where $Y^{t}$ is not available during training. We define a domain $\mathcal{D}$ as a pair consisting of a joint distribution $P$ on the input space and a labeling function $f:\Omega\rightarrow\mathcal{C}$. $P_{s}(X_{s}, Y_{s})$ and $P_{t}(X_{t},Y_{t})$ denote the joint distributions of the source and target domains, respectively. Our goal is to learn effective features such that the model also makes accurate predictions on the target domain.

\subsection{Bures joint Distribution Alignment Phase}

\begin{figure}[tb]
\centering\includegraphics[scale=0.52]{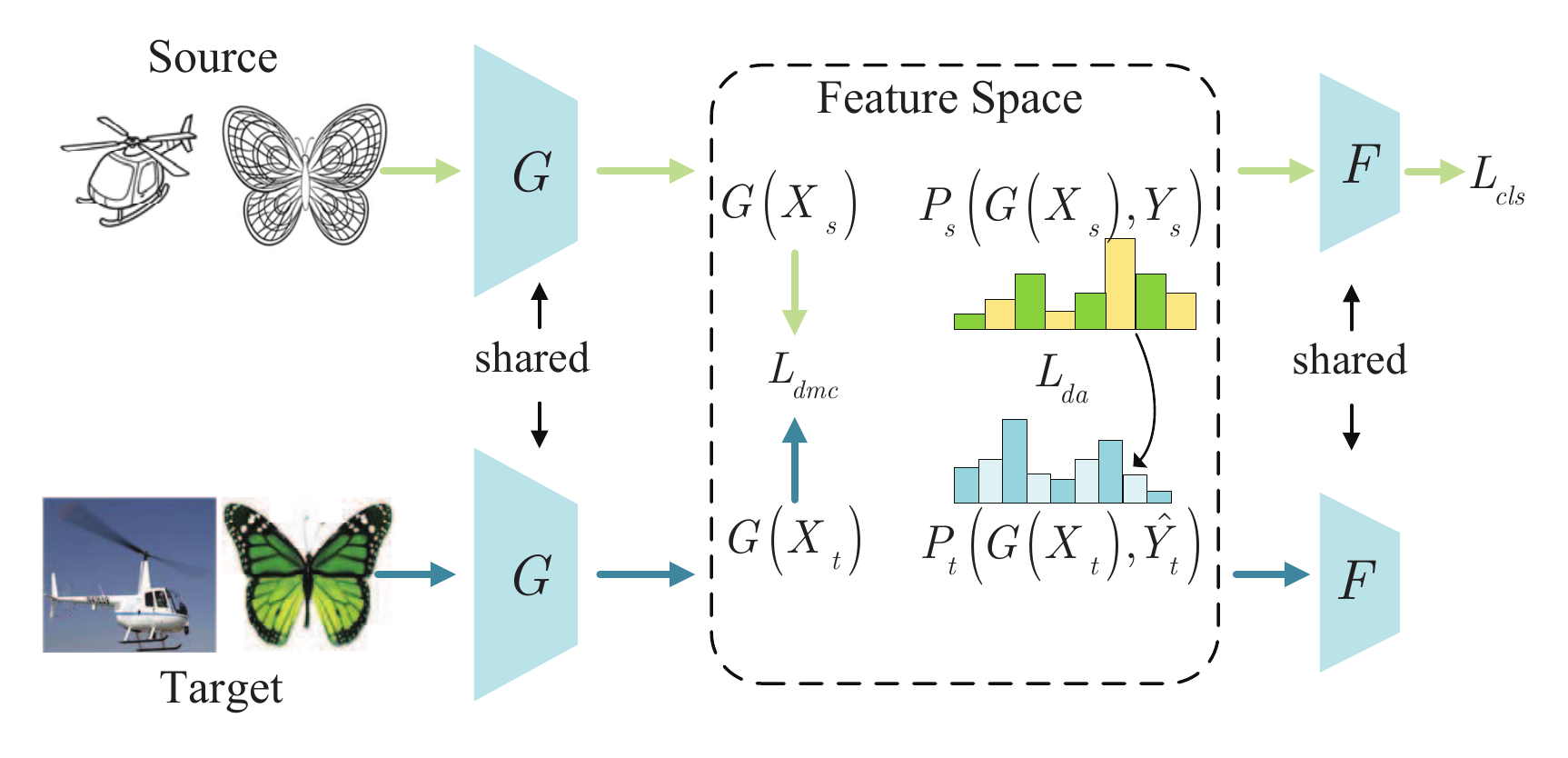}
\caption{Flowchart of the BJDA method. The feature extractor $G$ maps $X_{s}$ and $X_{t}$ into the feature space. The category classifier $F$ outputs the $C$-dimensional probability prediction vector for the input representation. $L_{da}$ computes the kernel Bures-Wasserstein distance between the joint distributions. $L_{dmc}$ improves the category discriminative ability of representations. $L_{cls}$ computes the cross-entropy loss for the source data. Better viewed in color.}\label{net1}
\end{figure}

Inspired by recent works about the joint distribution alignment for UDA \cite{JDOT2017, DeJDOT2018} as well as new proceedings \cite{OTrkhs} on the optimal transport theory in RKHS, we propose to align the joint distributions of the two domains in the infinite-dimensional kernel spaces. The architecture is depicted by Fig. \ref{net1}. Our model consists of a feature extractor $G$ with parameters $\theta_{G}$, a category classifier $F$ with parameters $\theta_{F}$. The feature extractor $G$ maps the samples to the representation space $\mathcal{G}$. $\mathcal{G}_{s}$ and $\mathcal{G}_{t}$ represent the subspace for $G(X_{s})$ and $G(X_{t})$, respectively. The category classifier $F$ outputs the $C$-dimensional probability prediction vector for the input representation. The optimal coupling between the joint distributions $P_{s}(G(X_{s}), Y_{s})$ and $P_{t}(G(X_{t}), Y_{t})$ is denoted as $\gamma^{\star}$, and we can get two sub-couplings $\gamma_{1}= \gamma^{\star} \big|_{\mathcal{G}_{s} \times \mathcal{G}_{t}}$ and $\gamma_{2}= \gamma^{\star} \big|_{\mathcal{C} \times \mathcal{C}}$. Since $Y_{t}$ is unavailable during training, we assign soft pseudo-labels $\hat{Y}_{t}$ to the unlabeled $X_{t}$, i.e., $\hat{Y}_{t}=F(G(X_{t}))$. We use the following theorems to model the discrepancy between the joint distributions of the two domains.

\textbf{Theorem 1 \cite{VilaniOT}.} Let ($\mathcal{Z}_{1},\mu$) and ($\mathcal{Z}_{2},\nu$) be two Polish probability spaces; let $a: \mathcal{Z}_{1} \rightarrow \mathbb{R} \cup\{-\infty\}$ and $b: \mathcal{Z}_{2} \rightarrow \mathbb{R} \cup\{-\infty\}$ be two upper semicontinuous functions such that $a \in L^{1}(\mu)$, $b \in L^{1}(\nu)$. Let $D: \mathcal{Z}_{1} \times \mathcal{Z}_{2} \rightarrow \mathbb{R} \cup\{+\infty\}$ be a lower semicontinuous cost function, such that $D(\mathbf{z}_{1}; \mathbf{z}_{2}) \geq a(\mathbf{z}_{1}) + b(\mathbf{z}_{2})$ for all $\mathbf{z}_{1} \in \mathcal{Z}_{1}$, $\mathbf{z}_{2} \in \mathcal{Z}_{2}$. Then there is a coupling of $(\mu,\nu)$ which minimizes the total cost $d_{W}(\mu,\nu)$ among all possible couplings $\gamma(\mathbf{z}_{1}; \mathbf{z}_{2})$.

\textbf{Theorem 2 \cite{VilaniOT}.} Any sub-coupling is still optimal.

Theorem 1 states the existence of an optimal coupling. Theorem 2 declares that any induced sub-coupling is also optimal.

In light of the above theorems, we can compute the squared Wasserstein distance between joint distributions as follows:
\begin{equation}
\begin{split}
\label{eq4kbw}
&d_{W}^{2}(P_{s}(G(X_{s}), Y_{s}),P_{t}(G(X_{t}), \hat{Y}_{t})) \\
=&\int D(\mathbf{g}_{s}, \mathbf{y}_{s} ; \mathbf{g}_{t}, \hat{\mathbf{y}}_{t}) d \gamma^{\star}(\mathbf{g}_{s}, \mathbf{y}_{s} ; \mathbf{g}_{t}, \hat{\mathbf{y}}_{t}) \\
=&\int D(\mathbf{g}_{s};\mathbf{g}_{t})d \gamma_{1}(\mathbf{g}_{s};\mathbf{g}_{t})+\int D(\mathbf{y}_{s} ; \hat{\mathbf{y}}_{t}) d \gamma_{2}(\mathbf{y}_{s} ; \hat{\mathbf{y}}_{t}),
\end{split}
\end{equation}
where $\mathbf{g}_{s} \in G(X_{s}),\mathbf{g}_{t} \in G(X_{t})$, $\mathbf{y}_{s} \in Y_{s}$, and $\hat{\mathbf{y}}_{t} \in \hat{Y}_{t}$.

The original OT problem is discussed in the Euclidean space. It is widely believed that the kernel methods are effective in exploring those complex data with nonlinear structures. Thus, the Bures-Wasserstein distance corresponding to the original OT problem is generalized to the infinite-dimensional settings, and the kernel Bures-Wasserstein distance \cite{OTrkhs} between the covariance operators in RKHS is induced.

Denote a nonempty set as $\mathcal{X}$, and let $\mathcal{H}$ be a Hilbert space of $\mathbb{R}$-valued function defined on $\mathcal{X}$. Let $k$ be a positive definite kernel on $\mathcal{X} \times \mathcal{X}$, and $\mathcal{H}_{\mathcal{X}}$ be the reproducing kernel Hilbert space generated by $k$. Define the implicit feature map $\phi:\mathcal{X}\rightarrow \mathcal{H_{\mathcal{X}}}$ as $\phi(\mathbf{x})=k(\cdot,\mathbf{x}), \forall \mathbf{x} \in \mathcal{X}$. Let $\mathbf{U}=[\mathbf{u}_{1},\cdots, \mathbf{u}_{m}]$ and $\mathbf{V}=[\mathbf{v}_{1},\cdots, \mathbf{v}_{n}]$ be two sample matrices from two probability measures $\mu$ and $\nu$, respectively. Let $\mathbf{\Phi}_{U}=[\phi(\mathbf{u}_{1}),\cdots, \phi(\mathbf{u}_{m})] $ and $\mathbf{\Phi}_{V}=[\phi(\mathbf{v}_{1}),\cdots, \phi(\mathbf{v}_{m})] $ be the mapped data matrices. Let $\mathbf{K}_{UU}$,$\mathbf{K}_{UV}$ and $\mathbf{K}_{VV}$ be the kernel matrices defined by $(\mathbf{K}_{UU})_{ij}=k(\mathbf{u}_{i},\mathbf{u}_{j})$,$(\mathbf{K}_{UV})_{ij}=k(\mathbf{u}_{i},\mathbf{v}_{j})$, and $(\mathbf{K}_{VV})_{ij}=k(\mathbf{v}_{i},\mathbf{v}_{j})$. Let $\mathbf{H}_{n}=\mathbf{I}_{n \times n} - \frac{1}{n}\mathbf{1}_{n}\mathbf{1}_{n}^{T}$ and  $\mathbf{H}_{m}=\mathbf{I}_{m \times m} - \frac{1}{m}\mathbf{1}_{m}\mathbf{1}_{m}^{T}$ be two centering matrices. The kernel Bures-Wasserstein distance \cite{OTrkhs} is expressed as:
\begin{equation}
\begin{split}
\label{eq3kbw}
\begin{aligned}
d_{BW}^{\mathcal{H}}(\mu,\nu)=&[\frac{1}{n} \operatorname{tr}(\mathbf{K}_{UU}\mathbf{H}_{n})+\frac{1}{m} \operatorname{tr}(\mathbf{K}_{VV}\mathbf{H}_{m}) \\
&-\frac{2}{\sqrt{m n}}\|\mathbf{H}_{n}\mathbf{K}_{UV}\mathbf{H}_{m}\|_{*}]^{\frac{1}{2}},
\end{aligned}
\end{split}
\end{equation}
where $\|\cdot\|_{*}$ denotes the nuclear norm, i.e., $\|\mathbf{A}\|_{*}=\sum_{i=1}^{r}\sigma_{i}(\mathbf{A})$, and $\sigma_{i}(\mathbf{A})$ is the singular values of matrix $\mathbf{A}$. This is an equivalent and computable formulation \cite{OTrkhs} of the kernel Bures-Wasserstein distance, the form of which is fully determined by the kernel function.

In order to exploit the capability of capturing nonlinear structures of kernel methods, we propose to match the joint distributions of source and target domains in RKHS. Specifically, we incorporate the formulation of the kernel Bures-Wasserstein distance and propose the following Bures joint distribution alignment loss
\begin{equation}
\begin{split}
\label{eq5}
L_{da}(\theta_{G}, \theta_{F})=&\frac{1}{n}\operatorname{tr}(\mathbf{K}_{g_{s}g_{s}}\mathbf{H}_{n})+\frac{1}{m}\operatorname{tr}(\mathbf{K}_{g_{t}g_{t}}\mathbf{H}_{m}) \\ &-\frac{2}{\sqrt{mn}}\|\mathbf{H}_{n}K_{g_{s}g_{t}}\mathbf{H}_{m}\|_{*}+\frac{1}{n}\operatorname{tr}(\mathbf{K}_{y_{s}y_{s}}\mathbf{H}_{n})\\
&+\frac{1}{m}\operatorname{tr}(\mathbf{K}_{\hat{y}_{t}\hat{y}_{t}}\mathbf{H}_{m})-\frac{2}{\sqrt{mn}}\|\mathbf{H}_{n}\mathbf{K}_{y_{s}\hat{y}_{t}}H_{m}\|_{*},
\end{split}
\end{equation}
where the kernel matrices follow the definition of Eq. (\ref{eq3kbw}), such as $(\mathbf{K}_{g_{s}g_{t}})_{ij}=k(\mathbf{g}_{i}^{s},\mathbf{g}_{j}^{t})$,$(\mathbf{K}_{y_{s}\hat{y}_{t}})_{ij}=k(\mathbf{y}_{i}^{s},\hat{\mathbf{y}}_{j}^{t})$. We use the Gauss kernel function in this paper, i.e., $k(\mathbf{g}_{i},\mathbf{g}_{j})=\exp(-\|\mathbf{g}_{i}-\mathbf{g}_{j}\|^{2}/\sigma^{2})$, where $\sigma^{2}$ is set as the mean of all the squared Euclidean distance $\|\mathbf{g}_{i}-\mathbf{g}_{j}\|^{2}$.

Our proposed Bures joint distribution alignment loss is differentiable by means of the closed form solutions, so we can directly model the joint distribution shift in the reproducing kernel Hilbert space. We also train the feature extractor $G$ and the category classifier $F$ with the cross-entropy loss on the source domain, i.e.,
\begin{equation}\label{eq6}
L_{cls}(\theta_{G}, \theta_{F})=-\mathbb{E}_{(x_{s},y_{s}) \sim P_{s}(X_{s},Y_{s})}[\mathbf{y}_{s}\log (F(G(\mathbf{x}_{s})))].
\end{equation}
\subsection{Contrastive Learning with Dynamic Margin}
\begin{figure}[htb]
\centering\includegraphics[scale=0.35]{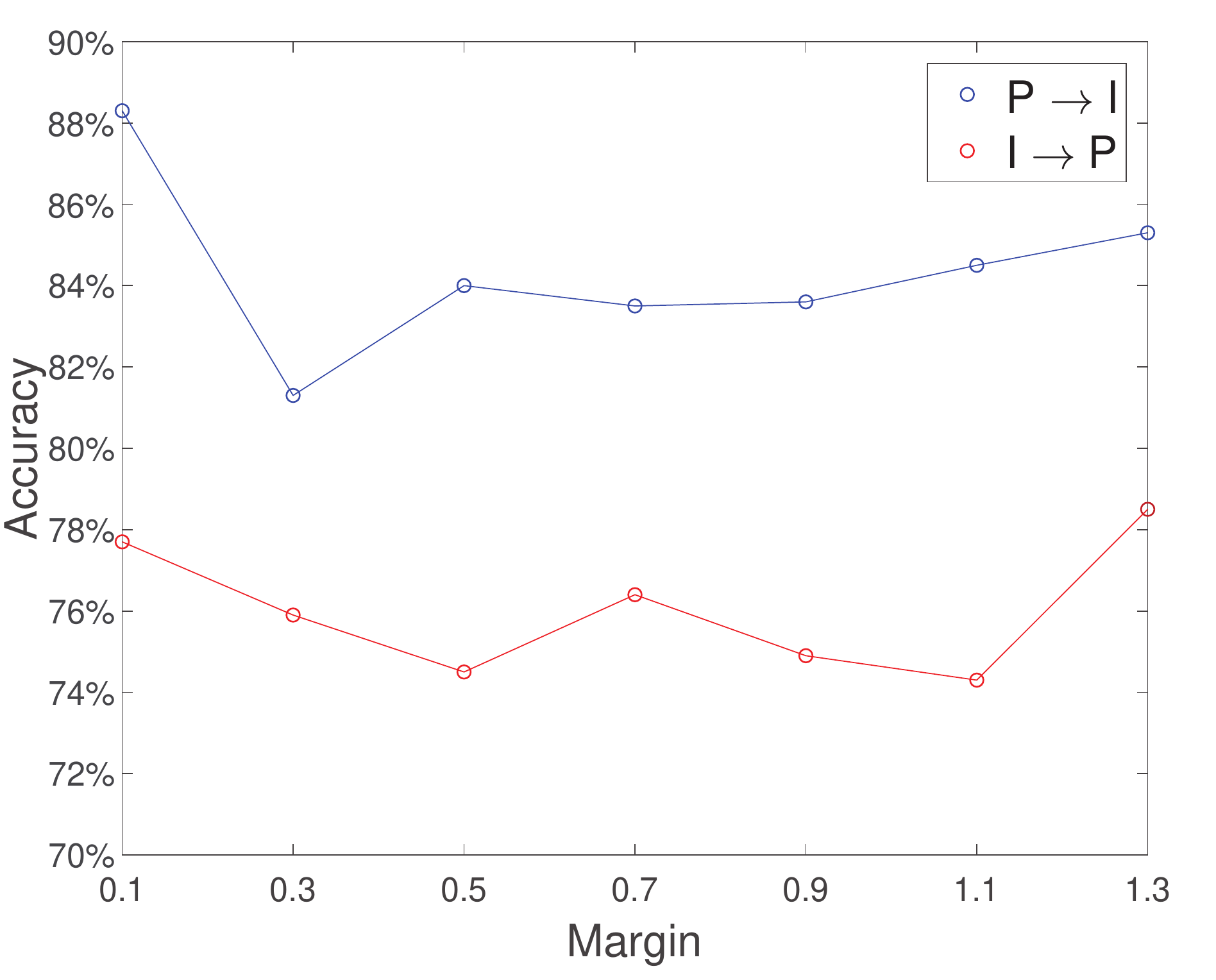}
\caption{Accuracy curves by different margin values of the original triplet loss. The tasks I$\rightarrow$P and P$\rightarrow$I in ImageCLEF-DA are used as examples. The accuracies show instability in response to different margin values. Better viewed in color.}\label{Galpha1}
\end{figure}

\begin{figure}[htb]
\centering\includegraphics[scale=0.65]{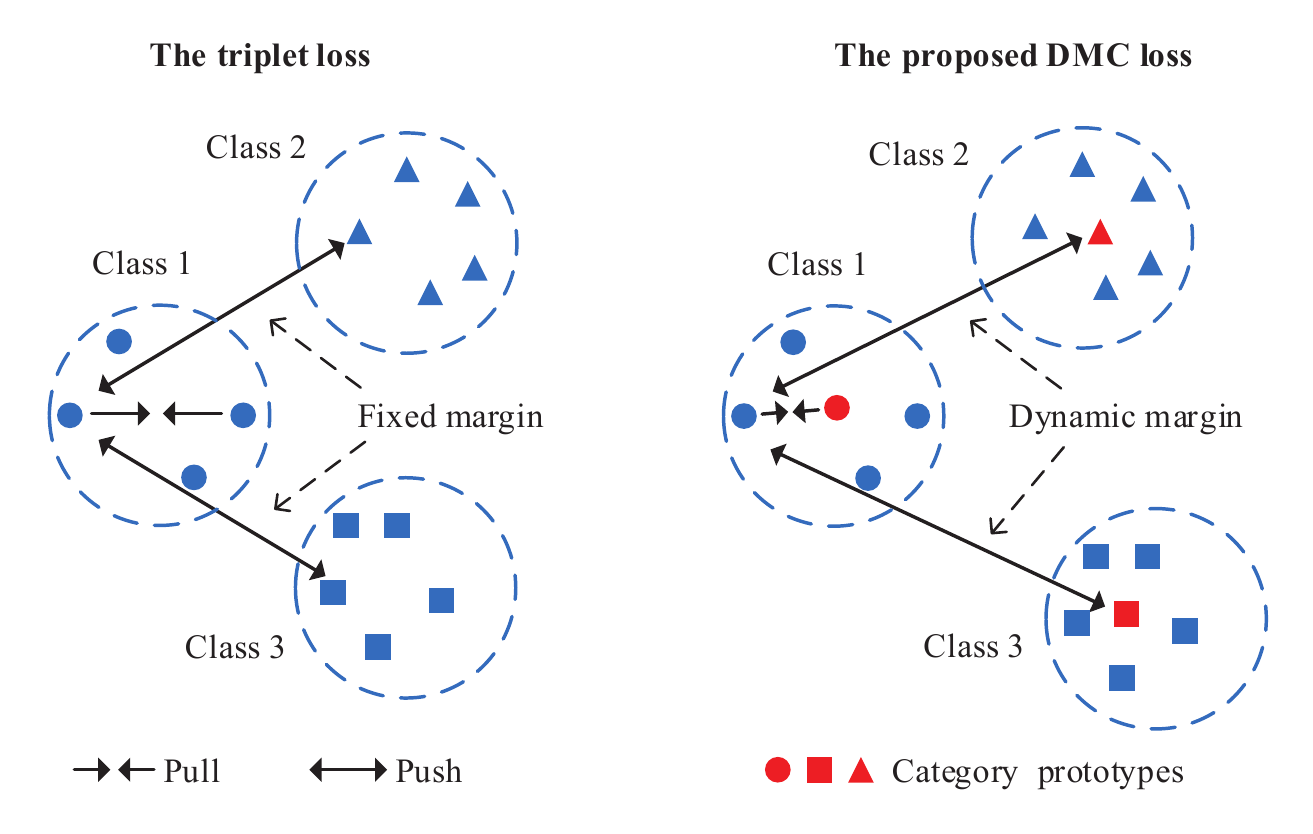}
\caption{Comparison between the original triplet loss and the proposed DMC loss. It is hard to train a feature space in which there exists fixed separation among various clusters based on the triplet loss. However, our DMC loss attempts to learn an dynamic separation between different clusters, and prompts the feature representations to become scattered around the corresponding category prototype. Better viewed in color.}\label{Gmac1}
\end{figure}

The goal of UDA is to train a model which makes accurate and robust predictions for the unlabeled target data. Therefore, enhancing the discriminative ability of the feature representations between categories may be of utmost importance. Recently, some works about contrastive learning are proposed. Wang et al.~\cite{Cross2021} use self-supervised learning method to reduce the discrepancy across domains. Park et al.~\cite{park2020joint} employ the mutual information between a feature and its label to train class-level discriminative features. Both~\cite{Cross2021} and~\cite{park2020joint} are based on the InfoNCE Loss~\cite{nce2018}, which minimizes the distance of all positive pairs relative to negative pairs, and they may probably encounter the potential risk of overfitting. Vikas et al.~\cite{Tow2021} propose a method for data augmentation, which mixes samples either at the input or hidden-state levels to create positive and negative samples. But those synthesis samples will introduce the Mixup-noise~\cite{Tow2021}. Intuitively, representations belonging to the same category should be located nearby in the representation space, and representations belonging to the different categories should be pushed far away. The triplet loss \cite{Trip2015,SRada,TripDA,ren19metric,huang21hyper} is effective in achieving the desired goal. Let $\mathbf{g}_{i}$ denote the feature representation of a sample $(\mathbf{x}_i,y_i)$, the triplet loss is defined as
\begin{equation}\label{eqTrip1}
L_{trip}(\theta_{G})=\sum_{i \neq j} \max\{\operatorname{Dist}(\mathbf{g}_{i},\mathbf{g}_{j}^{pos})-\operatorname{Dist}(\mathbf{g}_{i},\mathbf{g}_{j}^{neg})+ \alpha',0\},
\end{equation}
where $\alpha'$ represents a fixed margin value, $\mathbf{g}_{j}^{pos}$ denotes the feature representation that has the same category label with respect to $\mathbf{g}_{i}$, $\mathbf{g}_{j}^{neg}$ denotes the feature representation with different category label. We try to incorporate the triplet loss into our method, and train the network using the Bures joint distribution alignment loss, the cross-entropy loss, and the triplet loss. The experimental settings are described in Section \ref{section4}, and is termed as BJDA($w/o$ $L_{dmc}$)+$L_{trip}$. However, the experiment results show some limitations. As shown in Fig. \ref{Galpha1}, different margin values have a significant influence on the performance. It motivates us to find a dynamic margin to avoid the tedious task of determining the margin by cross-validation. On the other hand, the fixed margin may be an overly strict objective, as it is difficult to make all the clusters keep the same margin far away from each other in the high-dimensional representation space. In this work, we propose a dynamic margin based contrastive learning phase to address these issues and preserve the category discriminative information from the source domain.

Naturally, the prototypes, which play the role of each class center, contain much more robust category discriminative information than the sample points~\cite{ConvProto}. Therefore, we use the prototypes to guide the feature extractor to generate a stable and category discriminative feature space. Specifically, the representations that have the same label should get close to the corresponding category prototype and keep a reasonable distance far away from the category prototypes that have the different labels. Therefore, we propose to learn a dynamic margin among various clusters under the guidance of category prototypes and train a more robust and category-separable representation space.

Let $\mathbf{p}^{y_{i}}$ denote the prototype of the $y_{i}$-th category, $\mathbf{p}^{c}_{neg}$ is the prototype of the category $c$, where $c$ is different from $y_{i}$ and $c \in \mathcal{C}$. We propose the following dynamic margin contrastive (DMC) loss:
\begin{equation}
\begin{split}
\label{eq7}
&L_{dmc}(\theta_{G})=\\
&\sum_{i=1}^{n} \max\{\operatorname{Dist}(\mathbf{g}_{i},\mathbf{p}^{y_{i}})-\min_{c\neq y_{i}}[\operatorname{Dist}(\mathbf{g}_{i},\mathbf{p}^{c}_{neg})]+ \alpha,0\},
\end{split}
\end{equation}
where $\alpha=-\sum_{c=1}^{C}\hat{y}^{c}_{i}\log(\hat{y}^{c}_{i})$, $\hat{y}^{c}_{i}$ is the prediction probability of $F(\mathbf{g}_{i})$ for the category $c$, $\operatorname{Dist}(\cdot,\cdot)$ denotes the Euclidean distance between two feature vectors. The dynamic margin $\alpha$  will not be a fixed value, but has relation to the output of the network. When the margin $\alpha$ gets a small value, it means that the category classifier $F$ outputs a confident prediction for $\mathbf{g}_{i}$, and the feature representation $\mathbf{g}_{i}$ is far away from the decision categorical surface. It seems that $\mathbf{g}_{i}$ is a tractable sample for optimizing the representation space, thus we can assign a small margin between $\mathbf{p}^{y_{i}}$ and $\mathbf{p}^{c}_{Neg}$. On the contrary, when $\mathbf{g}_{i}$ gets an uncertain prediction, we should assign a larger margin between $\mathbf{p}^{y_{i}}$ and $\mathbf{p}^{c}_{Neg}$. The comparison between the original triplet loss and the proposed DMC loss is shown in Fig. \ref{Gmac1}. The triplet loss pulls the positive pairs (i.e., samples that have the same category label) close to each other and pushes the negative pairs (i.e., samples that have the different category label) far away from each other, but the margin is a fixed value which is hard to determine. Our DMC loss encourages the feature representations scattering around the center of the corresponding category and seeks to keep a reasonable separation between different clusters.

The prototypes are computed from the source data which have ground truth labels. To generalize the discriminative information from the source domain to the target domain, we compute the DMC loss using both the source and target data. Since labels of the target data are unavailable during training, we assign hard pseudo-labels $\hat{Y}_{t}^{'}$ to the unlabeled $X_{t}$, that is,
\begin{equation}\label{eq7hard}
\hat{Y}_{t}^{'}=\mathop{\arg\max}\limits_{i}{\{[F(G(X_{t}))]_{i}\}}_{i=1}^{C}.
\end{equation}

Although both our DMC and previous works~\cite{Cross2021,park2020joint,Tow2021} are based on the idea of contrastive learning, their models are different. Specifically, the works~\cite{Cross2021,park2020joint,Tow2021} search positive samples and negative samples for each sample and compute the distance for all positive pairs and negative pairs, and they assign a fixed margin between clusters. While our DMC only needs to compute the distance between a single sample and two category prototypes, and it introduces a dynamic margin to flexibly characterize the class separability of deep features. Overall, the proposed DMC loss encourages maintaining a reasonable separation across the centers of feature representations belonging to different categories in the feature space. Since the relationship of intra-category compactness and inter-category discrepancy are taken into account, the feature learned for the target data has the better discriminative ability.

\subsection{Model Training}
The objective function of the BJDA method is formulated as
\begin{equation}\label{eq8}
\min_{\theta_{G}, \theta_{F}} L_{cls}(\theta_{G}, \theta_{F})+ \lambda_1 L_{da}(\theta_{G}, \theta_{F}) + \lambda_2 L_{}(\theta_{G}),
\end{equation}
where $\lambda_1$ and $\lambda_2$ are non-negative hyper-parameters.

We use the mini-batch stochastic gradient descent (Mini-batch SGD)~\cite{miniSGD2014} to solve the objective function.

The detailed training procedure is shown in $\textbf{Algorithm 1}$.
\begin{algorithm}[ht]
	\caption{Pseudo-code for \textbf{BJDA} algorithm.}
    \textbf{Input}: {Labeled source data $(X_{s},Y_{s})$, unlabeled target data $X_{t}$, maximum iteration $T_{max}$, parameters $\lambda_1$ and $\lambda_2$.}

    \textbf{Output}: {Feature extractor $G$, category classifier $F$.}

    \begin{algorithmic}[1]
		\STATE Initialize the feature extractor $G$ and the category classifier $F$ with xavier\cite{xavier2010};
        \FOR{$i=1$ to $T_{max}$}
        \STATE{Randomly sample a mini-batch of data from $(X_{s},Y_{s})$ and $X_{t}$;}
        \STATE{Using the feature extractor $G$ to learn feature representations for the data;}
        \STATE{Compute the category prototypes with $(G(X_{s}),Y_{s})$;}
        \STATE{Feed the feature representations into the category classifier $F$ and calculate the cross-entropy loss as Eq. (\ref{eq6});}
        \STATE{Get the pseudo labels $\hat{Y}_{t}$ and $\hat{Y}_{t}^{'}$ for $X_{t}$;}
        \STATE{Calculate the Bures joint distribution alignment loss as Eq. (\ref{eq5});}
        \STATE{Calculate the dynamic margin contrastive loss as Eq. (\ref{eq7});}
        \STATE{Update the feature extractor $G$ and category classifier $F$ by Mini-batch SGD;}
        \ENDFOR
	\end{algorithmic}
\end{algorithm}

\section{Experimental Results}\label{section4}

In this section, we first evaluate our BJDA method on the image classification tasks under UDA settings. Next, we perform ablation study and convergence analysis on BJDA. Then we show the impact of different hyper-parameters.
\subsection{Datasets}

We use six benchmark datasets, i.e., ImageCLEF-DA, Office-Caltech, Adaptiope, and Refurbished Office-31.

\textbf{ImageCLEF-DA:} The ImageCLEF-DA dataset \cite{Long2017JAN} is a commonly used dataset for UDA. It consists of images from three distinct domains, which are subsets of Caltech-256 ($C$), ImageNet ILSVRC 2012 ($I$), and Pascal VOC 2012 ($P$) datasets. It selects 12 classes common to the three public datasets. There are 50 images in each category, and each domain is comprised of the same amount of samples.

\textbf{Office-Caltech10:} The Office-Caltech10 dataset \cite{GFK2012cvpr1} is a dataset widely adopted by UDA methods. The whole dataset contains 2533 images, which consists of 4 domains: Amazon (images downloaded from online websites), DSLR (high-resolution images from digital SLR cameras), Webcam (low-resolution images taken by web cameras), and Caltech-10 (images from Caltech-256 dataset). We use characters $A$, $D$, $W$, $C$ to represent these domains for short, respectively. There are 10 common categories of objects in each domain, including monitor, keyboard, headphones, calculator, etc.

\begin{figure}[htb]
\centering\includegraphics[scale=0.6]{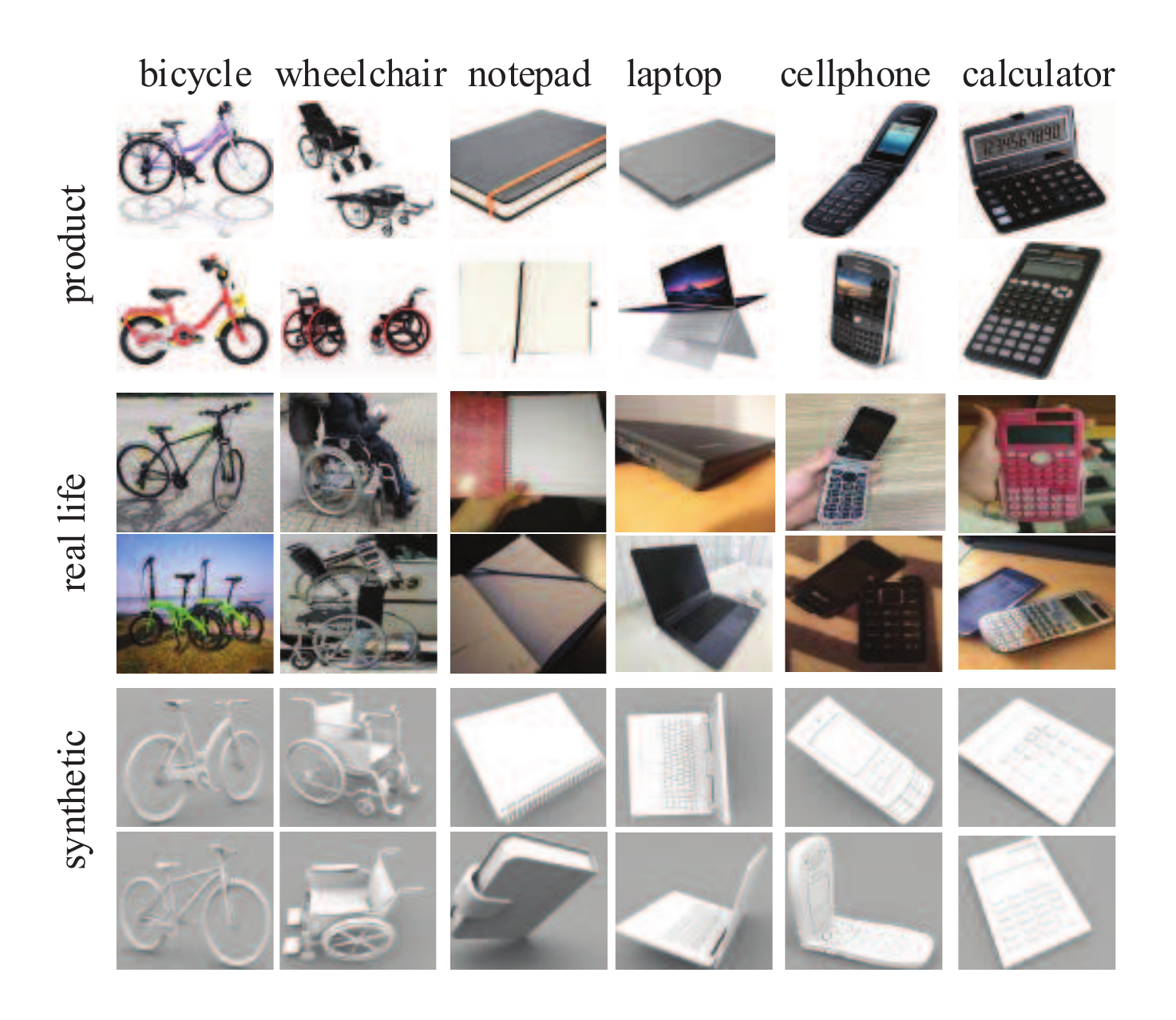}
\caption{Example images from the recently proposed Adaptiope dataset. From top to bottom: product, real life and synthetic domain. The column-wise images come from different categories. Images from different domains present diverse characteristics. Better viewed in color.}\label{Adapts4}
\end{figure}

\textbf{Adaptiope:} The Adaptiope dataset \cite{Ref31} is a new, large, diverse classification dataset for UDA. It has 3 domains, namely Product, Real life, and Synthetic. Images for the Product domain are real article images from the e-commerce website, and images in the Real life domain are customer review images. For the Synthetic domain, data is collected from artistic websites such as $sketchfab.com$. It is noteworthy that every category has an equal number of images, and each domain contains 12300 images from 123 classes. Fig. \ref{Adapts4} presents example images from the Adaptiope dataset. It can be seen that there are large domain gaps among various domains in this challenging dataset, and different backgrounds and colorspaces exist among images from different domains.

\textbf{Refurbished Office-31:} The Refurbished Office-31 dataset\cite{Ref31} is an upgrade version of the commonly used Office31 dataset \cite{Saenko2010da1}. In light of there are some systematic errors (i.e., ambiguous ground truth, domain leakage) in the Amazon domain of the original Office31 dataset \cite{Saenko2010da1}, the author of \cite{Ref31} propose the Refurbished Office-31 dataset by manually cleaning a total of 834 of the 2,817 images in the Amazon domain. It is a collection of images from three different domains, i.e., The Refurbished Amazon, DSLR, and Webcam. We use $A_{ref}$, $D$, $W$ to represent these domains for short, respectively.

\textbf{Office-Home:} It is one of the largest datasets for the UDA tasks in the community~\cite{HOME2017}. It has $15, 500$ samples with 65 categories. It consists of four different domains, namely Art, Clipart, Product, Real-World. We use $A$, $C$, $P$, $R$ to represent these domains for short, respectively.

\textbf{DomainNet:} It is a recently proposed benchmark~\cite{Peng_2019_ICCV} with 0.59 million images, it consists of six different domains, namely Clipart, Infograph, Painting, Quickdraw, Real, and Sketch\cite{Peng_2019_ICCV}. Each domain contains 345 categories of objects. We use $clp$, $inf$, $pnt$, $qdr$, $rel$, $skt$ to represent these domains for short, respectively.

We compare our BJDA method against a variety of state-of-the-art UDA methods, including CORAL \cite{coral2016AAAI}, DANN \cite{Yaroslav2016JMLR}, DAN \cite{Long2019PAMI1}, OT-GL \cite{OTuda1}, CDAN \cite{cdan1}, SymNets \cite{TanMK2019}, ATM \cite{Atm2021}, RSDA \cite{Rsda2020}. Results of these methods are quoted from the original papers, or re-implemented under the same UDA settings. In the test phase, test samples are forwarded through the feature extractor $G$ and the category classifier $F$. The setting of Source-only is used as a baseline, where all images from source domains are used to train a classifier without matching with the target data.

\subsection{Experimental Settings}
To make it a fair comparison, we follow the experimental settings of UDA (\cite{TanMK2019, Long2019PAMI1, Atm2021} ). We perform each task for five runs and report the average accuracy and standard deviation. We use the deep features commonly used in existing works for a fair comparison with other UDA methods. Specifically, the Decaf6 \cite{decaf2014} features (activations of the 6\textit{th} fully connected layer of the convolutional neural network, the dimensionality is 4096) are used for the Office-Caltech10 dataset, the ResNet101~\cite{resnet2016}features (activations of the penultimate fully connected layer of the convolutional neural network, the dimensionality is 2048) are used for the DomainNet dataset. In light of ResNet50~\cite{resnet2016} has been commonly used to extract features or as the backbone of deep learning models in the literature, we use the ResNet50 features (the dimensionality is 2048) for the other three datasets in our experiments. For the feature extractor $G$, we use two-layered fully-connected networks with 1024 neurons in the hidden layers, in which the LeakyReLU is used as the activation function. We set the number of output dimensions of the feature extractor as 512. For the category classifier $F$, we use a fully-connected network. The maximum iteration $T_{max}$ is set to 300 in all experiments. All the networks are optimized by a Minibatch SGD optimizer with a weight decay of 0.0005 and momentum of 0.9, and the initial learning rate is 0.001. The values of $\lambda_1$ and $\lambda_2$ for experiments in current manuscript are set as (0.5,0.3). We follow the protocol in the community~\cite{Long2017JAN, Long2019PAMI1, Atm2021} and tune hyper-parameters using transfer cross-validation~\cite{CV2010}. All the experiments are implemented using the PyTorch library and are trained with an NVIDIA GeForce TITAN Xp GPU.
\subsection{Experimental Results on ImageCLEF-DA}
\begin{table*}[tbp]
  \centering
  \caption{Accuracy (\%) comparison between BJDA and other UDA methods on ImageCLEF-DA (ResNet50).}
    \begin{tabular}{cccccccc}
    \hline
    Method & I$\rightarrow$P  & P$\rightarrow$I  & I$\rightarrow$C  & C$\rightarrow$I  & C$\rightarrow$P  & P$\rightarrow$C  & Avg \\
    \hline
    Source-only &74.8$\pm$0.3  & 83.9$\pm$0.1  & 91.5$\pm$0.3  & 78.0$\pm$0.2  & 65.5$\pm$0.3  & 91.2$\pm$0.3  & 80.7 \\
    \hline
    CORAL~\cite{coral2016AAAI} &76.9$\pm$0.2  & 88.5$\pm$0.3  & 93.6$\pm$0.3  & 86.8$\pm$0.6  & 74.0$\pm$0.3  & 91.6$\pm$0.3 & 85.2 \\
    \hline
    JAN~\cite{Long2017JAN}& 76.8&88.0&94.7&89.5&74.2&91.7&85.8 \\
    \hline
     DDC~\cite{DDC2014} &74.6$\pm$0.3  & 85.7$\pm$0.8  & 91.1$\pm$0.3  & 82.3$\pm$0.7  & 68.3$\pm$0.4  & 88.8$\pm$0.2  & 81.8 \\
    \hline
    DANN~\cite{Yaroslav2016JMLR} &75.0$\pm$0.6  & 86.0$\pm$0.3  & 96.2$\pm$0.4  & 87.0$\pm$0.5  & 74.3$\pm$0.5  & 91.5$\pm$0.6  & 85.0 \\
    \hline
    DAN~\cite{Long2019PAMI1} &75.0$\pm$0.4  & 86.2$\pm$0.2  & 93.3$\pm$0.2  & 84.1$\pm$0.4 & 69.8$\pm$0.4  & 91.3$\pm$0.4  & 83.3 \\
    \hline
    OT-GL~\cite{OTuda1} & 73.2&83.1&93.6&85.5&71.3&91.7&83.1 \\
    \hline
    DeepJDOT~\cite{DeJDOT2018} &  79.4&86.7&93.2&82.0&69.5&91.3&83.7 \\
    \hline
    MADA~\cite{pei2018multi} &75.0$\pm$0.3  & 87.9$\pm$0.2  & 96.0$\pm$0.3  & 88.8$\pm$0.3  & 75.2$\pm$0.2  & 92.2$\pm$0.3  & 85.8 \\
    \hline
    CDAN~\cite{cdan1} & 77.7$\pm$0.3  & 90.7$\pm$0.2  & 97.7$\pm$0.3  & 91.3$\pm$0.3  & 74.2$\pm$0.2  & 94.3$\pm$0.3  & 87.7 \\
    \hline
    SymNets~\cite{TanMK2019} & 80.2$\pm$0.3  & 93.6$\pm$0.2  & 97.0$\pm$0.3  & 93.4$\pm$0.3  & 78.7$\pm$0.2  & 96.4$\pm$0.1  & 89.9 \\
    \hline
    RSDA~\cite{Rsda2020} & 79.2$\pm$0.4  & 93.0$\pm$0.2  & 98.3$\pm$0.4  & 93.6$\pm$0.4  & 78.5$\pm$0.3  & \textbf{98.2$\pm$0.2}  & 90.1 \\
    \hline
    ATM~\cite{Atm2021} & 80.3$\pm$0.3  & 92.9$\pm$0.4  & \textbf{98.6$\pm$0.4}  & 93.5$\pm$0.1  & 77.8$\pm$0.3  & 96.7$\pm$0.2  & 90.0 \\
    \hline
    TCM~\cite{TCM2021} & 79.9$\pm$0.4  & \textbf{94.2$\pm$0.2}  & 97.8$\pm$0.3  & 93.8$\pm$0.4  & 79.9$\pm$0.4  & 96.9$\pm$0.4  & 90.5 \\
    \hline
    BJDA (ours) & \textbf{81.9$\pm$0.4}  &93.8$\pm$0.1  & 97.5$\pm$0.4  & \textbf{95.8$\pm$0.2}  & \textbf{81.9$\pm$0.1}  & 96.3$\pm$0.1  & \textbf{91.2} \\
    \hline
    \end{tabular}%
  \label{Clef}%
\end{table*}
The classification results for the ImageCLEF-DA dataset are reported in Table \ref{Clef}. The results show that our BJDA significantly outperforms all methods on 3 out of 6 UDA tasks and achieves the best average accuracy. There are several interesting observations as shown in Table \ref{Clef}. Firstly, BJDA achieves better performance than DeepJDOT \cite{DeJDOT2018} which matches the joint distributions using OT and deep neural networks in the Euclidean space. This confirms that performing domain alignment via OT in RKHS can improve the performance of UDA tasks. Secondly, several methods (i.e., MADA \cite{pei2018multi}, CDAN \cite{cdan1}, SymNets \cite{TanMK2019}, RSDA \cite{Rsda2020}, ATM \cite{Atm2021}, TCM~\cite{TCM2021}) that explicitly or implicitly consider the label information significantly outperform those methods that perform domain alignment with only feature marginal distribution (i.e., DDC \cite{DDC2014},DANN \cite{Yaroslav2016JMLR}, DAN \cite{Long2019PAMI1}, OT-GL \cite{OTuda1}), which indicates that label information contributes to modeling the multi-modal structure of image data. Thirdly, our BJDA substantially improves the classification accuracy of difficult UDA tasks (e.g., I$\rightarrow$P, C$\rightarrow$I, C$\rightarrow$P), where there exist large domain gaps, which confirms that aligning the joint distributions among various domains is robust to large domain gaps. Finally, our BJDA achieves the best average performance, which demonstrates its effectiveness in modeling a more transferable target classifier.

\subsection{Experimental Results on Office-Caltech10}
\begin{table*}[tbp]
  \centering
  \caption{Accuracy (\%) comparison between BJDA and other UDA methods on Office-Caltech10 (Decaf6).}
  \renewcommand{\tabcolsep}{0.25pc} 
  \renewcommand{\arraystretch}{1.1} 
    {
    \begin{tabular}{cccccccccccccc}
    \hline
Method & A$\rightarrow$C  & A$\rightarrow$D  & A$\rightarrow$W  & C$\rightarrow$A  & C$\rightarrow$D  & C$\rightarrow$W & D$\rightarrow$A  & D$\rightarrow$C & D$\rightarrow$W  & W$\rightarrow$A & W$\rightarrow$C  & W$\rightarrow$D & Avg \\
     \hline
     Source-only &  82.7  & 85.4 & 78.3 & 91.5  & 88.5 & 83.1 & 80.6 & 74.6 & 99.0  & 77.0  & 69.6.7  & 100.0  & 84.2 \\
     \hline
     CORAL \cite{coral2016AAAI}&  85.3  & 80.8  & 76.3   & 91.1  & 86.6  & 81.1 & 88.7  & 80.4  & 99.3  & 82.1  & 78.7  & 100.0  & 85.9 \\
     \hline
     JDA \cite{JDA2013}&  81.3  & 86.3  & 80.3   & 88.0  & 84.1  & 79.6 & 91.3  & 81.1  & 97.5  & 90.2  & 82.0  & 98.9  & 86.7 \\
     \hline
     DDC \cite{DDC2014}&  85.0&89.0&86.1&91.9&88.8&85.4&89.5&81.1&98.2&84.9&78.0&100.0&88.2 \\
     \hline
     DANN \cite{Yaroslav2016JMLR}&  87.8&82.5&77.8&93.3&91.2&89.5&84.7&82.1&98.9&82.9&81.3&100.0& 87.7 \\
     \hline
     DAN \cite{Long2019PAMI1} &84.1&91.7&91.8&92.0&89.3&\textbf{90.6}&90.0&80.3&98.5&92.1&81.2&100.0&90.1 \\
     \hline
     OT-GL \cite{OTuda1}& 85.5  & 85.0  & 83.1   & 92.1  & 87.3  & 84.2 & 92.3  & 84.1  & 96.3  & 90.6  & 81.5  & 96.3  & 88.2 \\
     \hline
     DeepJDOT \cite{DeJDOT2018}&  87.4  & 88.5  & 86.7   & 92.3  & 92.0  & 85.3 & 91.5  & 85.3  & 98.7  & 86.6  & 84.7  & 98.7  & 89.8 \\
     \hline
     DMP \cite{YouWeiPami1}&  86.6  & 90.4  & 91.3   & 92.8  & 93.0  & 88.5 & 91.4  & 85.3  & 97.7  & 91.9  & 85.6  & 100  & 91.2 \\
     \hline
     BJDA (ours)&  \textbf{87.8}& \textbf{92.9}  & \textbf{94.2}   & \textbf{93.5}  & \textbf{94.3}  & 90.5 & \textbf{92.7}  & \textbf{86.0}  & \textbf{100.0}  & \textbf{93.3}  & \textbf{86.3}  & \textbf{100.0}  & \textbf{92.6} \\
    \hline
    \end{tabular}}
    \label{Ofc10}
\end{table*}
Table~\ref{Ofc10} summarizes the classification results for the Office-Caltech10 dataset. The BJDA model substantially outperforms the comparison methods on most UDA tasks. Specifically, BJDA obtains an absolute accuracy improvement of 1.4\% over the latest competitor DMP \cite{YouWeiPami1}. Notably, BJDA and DAN \cite{Long2019PAMI1} yield greater improvements for this benchmark, which highlights the advantage of performing distribution alignment by exploiting the nonlinear feature mapping of RKHS. For JDA \cite{JDA2013} and DeepJDOT \cite{DeJDOT2018}, both of which match the joint distribution cross the two domains, BJDA outperforms them by 5.9\% and 2.8\%, respectively. It validates the the superiority of BJDA which directly models the joint distribution shift in the infinite-dimensional spaces. UDA tasks with the domain A are more challenging as images from domain A exhibit more diversity within each category. In this case, aligning the joint distributions can effectively improve the performance of UDA tasks. As Table \ref{Ofc10} shows, our BJDA method exhibits significant advantages over other UDA methods when domain $A$ acts as the source domain or the target domain. The results show that BJDA tends to more stable when there are large domain shift between the source and target domains.
\subsection{Experimental Results on Adaptiope}
\begin{table*}[tbp]
  \centering
  \caption{Accuracy (\%) comparison between BJDA and other UDA methods on Adaptiope (ResNet50).}
    \begin{tabular}{cccccccc}
    \hline
    Method & P$\rightarrow$R & P$\rightarrow$ S & R$\rightarrow$P & R$\rightarrow$S & S$\rightarrow$P  & S$\rightarrow$R  & Avg \\
    \hline
    Source-only & 63.6$\pm$0.4 & 26.7$\pm$1.7  & 85.3$\pm$0.3   & 27.6$\pm$0.3  & 43.1$\pm$0.3  & 29.6$\pm$0.4  & 45.9 \\
    \hline
     CORAL \cite{coral2016AAAI}& 67.5  & 38.2 & 86.7 & 29.1 & 45.3 & 35.2  & 50.3 \\
     \hline
     DDC \cite{DDC2014}&  69.5  & 36.2  & 87.5 & 29.4 & 43.5 & 33.5  & 49.9  \\
    \hline
    DANN \cite{Yaroslav2016JMLR} &  68.5 & 39.3 & 89.2 & 37.4  & 53.7  & 35.1  & 53.8 \\
    \hline
    OT-GL \cite{OTuda1}& 66.1 & 42.2 & 87.4  & 37.6  & 53.8 & 40.9 & 54.7 \\
    \hline
    DeepJDOT \cite{DeJDOT2018} & 68.8 & 43.9 & 86.5   & 40.1 & 55.3 & 43.5 & 56.4\\
    \hline
    SymNets~\cite{TanMK2019} & \textbf{81.4$\pm$0.3}  & 53.1$\pm$0.6  & \textbf{92.3$\pm$0.2}   & 49.2$\pm$1.0  & 69.6$\pm$0.5  & 44.9$\pm$1.5  & 65.1 \\
    \hline
     ATM \cite{Atm2021} & 73.9 & 49.6 & 87.0   & 45.9 & 61.3 & 35.3 & 58.8\\
    \hline
    RSDA~\cite{Rsda2020} & 73.8$\pm$0.2   & 59.2$\pm$0.2 & 87.5$\pm$0.2   & 50.3$\pm$0.5  & 69.5$\pm$0.6  & 44.6$\pm$1.1  & 64.2 \\
    \hline
    BJDA (ours)& 74.3$\pm$0.1 & \textbf{60.1$\pm$0.4} & 88.4$\pm$0.2 & \textbf{52.0$\pm$0.2} & \textbf{76.4$\pm$0.2}  & \textbf{56.3$\pm$0.1} & \textbf{67.9} \\
    \hline
    \end{tabular}%
  \label{Adapt1}%
\end{table*}
To demonstrate the applicability of the proposed BJDA on more intricate real-world problems, we evaluate BJDA on the recently proposed Adaptiope benchmark. The results compared with several state-of-the-art methods are shown in Table \ref{Adapt1}. Our method outperforms the other methods on 4 out of 6 UDA tasks and achieves the best average accuracy. Especially in the most challenging task (i.e, P$\rightarrow$ S, R$\rightarrow$S, and S$\rightarrow$R), for which Source-only accuracies are below 30\%, our method significantly outperforms all the previous methods. It shows the superiority of our BJDA when there are severe domain gaps between the source and the target domains. Furthermore, both BJDA and DeepJDOT \cite{DeJDOT2018} show significant improvements over the marginal-based methods (i.e., CORAL \cite{coral2016AAAI}, DDC \cite{DDC2014}, DANN \cite{Yaroslav2016JMLR}, OT-GL \cite{OTuda1}), which demonstrates that it is very necessary to align the joint distributions rather than the marginal distributions. Our BJDA achieves 2.8\% and 3.7\% average accuracy improvements with respect to the recent SymNets \cite{TanMK2019} and RSDA \cite{Rsda2020}, respectively. These experimental results indicate that BJDA can extract more transferable features than state-of-the-art approaches.

\subsection{Experimental Results on Refurbished Office-31}
\begin{table*}[tbp]
  \centering
  \caption{Accuracy (\%) comparison between BJDA and other UDA methods on Refurbished Office-31 (ResNet50).}
    \begin{tabular}{cccccccc}
    \hline
    Method & A$_{ref}$$\rightarrow$D &A$_{ref}$$\rightarrow$ W & D$\rightarrow$A$_{ref}$ & D$\rightarrow$W & W$\rightarrow$A$_{ref}$  & W$\rightarrow$D  & Avg \\
    \hline
    Source-only & 79.2$\pm$0.6 & 76.8$\pm$1.0  & 73.5$\pm$1.2   & 96.3$\pm$0.2  & 74.1$\pm$0.5  & 99.1$\pm$0.3  & 83.2 \\
    \hline
     CORAL \cite{coral2016AAAI}&  84.7  & 81.5  & 78.8 & 97.2 & 79.5 & 99.6  & 86.8 \\
     \hline
     DDC \cite{DDC2014}&  85.3  & 82.0  & 78.1 & 97.7 & 77.8 & 98.9  & 86.6  \\
    \hline
    DANN \cite{Yaroslav2016JMLR} &  86.4   & 87.4  & 75.9   & 97.9  & 76.1  & 99.1  & 87.1 \\
    \hline
    OT-GL \cite{OTuda1}&  80.1   & 80.9  & 81.1   & 96.5  & 78.9  & 96.7  & 85.7 \\
    \hline
    DeepJDOT \cite{DeJDOT2018} &  82.2   & 80.1  & 82.2   & 96.2  & 82.2  & 99.2  & 86.0\\
    \hline
    SymNets~\cite{TanMK2019} & 92.4$\pm$0.4  & 91.0$\pm$0.2  & 90.6$\pm$0.4   & 98.0$\pm$0.1  & \textbf{89.2$\pm$0.4}  & 99.8$\pm$0.0  & 93.5 \\
    \hline
     ATM \cite{Atm2021} &  88.5   & 91.7  & 72.4   & 99.2  & 76.6  & 99.2  & 88.1\\
    \hline
    RSDA~\cite{Rsda2020} & 90.9$\pm$1.3    & \textbf{91.8$\pm$0.5}  & 87.3$\pm$0.6   & 98.8$\pm$0.2  & 90.5$\pm$0.9  & \textbf{99.9$\pm$0.1}  & 93.2 \\
    \hline
    BJDA (ours) & \textbf{92.8$\pm$0.1} & 91.3$\pm$0.1 & \textbf{92.5$\pm$0.2} & \textbf{99.3$\pm$0.1} & 87.9$\pm$0.1   & 99.8$\pm$0.2 & \textbf{93.9} \\
    \hline
    \end{tabular}%
  \label{Ref31}%
\end{table*}
We report the results of the proposed BJDA method on the recently proposed Refurbished Office-31 benchmark in Table~\ref{Ref31}. It can be seen that our BJDA method outperforms all comparison methods on 3 out of 6 UDA tasks and achieves the best average results. It is interesting to see that BJDA, SymNets \cite{TanMK2019}, RSDA \cite{Rsda2020} achieve significantly better results over individual UDA tasks than the other comparison methods, the reason may be that these methods take into account the label information and lead to a more discriminative target feature space. By comparing the performance of our BJDA with the DeepJDOT \cite{DeJDOT2018} which also considers aligning the joint distribution based on OT, we find that BJDA exceeds DeepJDOT on all UDA tasks by large margins. The results verify that it is effectual to extend the OT theory in the Euclidean space to the reproducing kernel Hilbert space. It is worth noting that our BJDA achieves a significant improvement on the hard tasks between the domain A$_{ref}$ and the domain D, which demonstrates the effectiveness of our proposed approach.
\subsection{Experimental Results on Large-scale Datasets}
\begin{table*}[tbp]
  \centering
  \caption{Accuracy (\%) comparison between BJDA and other UDA methods on Office-Home (ResNet50).}
  \renewcommand{\tabcolsep}{0.25pc} 
  \renewcommand{\arraystretch}{1.1} 
    {
    \begin{tabular}{cccccccccccccc}
    \hline
Method & A$\rightarrow$C & A$\rightarrow$P & A$\rightarrow$R & C$\rightarrow$A & C$\rightarrow$P & C$\rightarrow$R & P$\rightarrow$A & P$\rightarrow$C & P$\rightarrow$R & R$\rightarrow$A & R$\rightarrow$C & R$\rightarrow$P & Avg \\
     \hline
     Source-only &  34.9 & 50.0 & 58.0 & 37.4 & 41.9 & 46.2 & 38.5 & 31.2 & 60.4 & 53.9 & 41.2 & 59.9 & 46.1 \\
     \hline
    JAN \cite{Long2017JAN} &45.9 & 61.2 & 68.9 & 50.4 & 59.7 & 61.0 & 45.8 & 43.4 & 70.3 & 63.9 & 52.4 & 76.8 & 58.3 \\
     \hline
     DANN \cite{Yaroslav2016JMLR}&  45.6 & 59.3 & 70.1 & 47.0 & 58.5 & 60.9 & 46.1 & 43.7 & 68.5 & 63.2 & 51.8 & 76.8 & 57.6 \\
     \hline
     DAN \cite{Long2019PAMI1} &43.6 & 57.0 & 67.9  &  45.8 & 56.5 & 60.4  & 44.0  & 43.6 & 67.7 & 63.1 & 51.5 & 74.3 & 56.3 \\
     \hline
     KGOT \cite{OTrkhs}& 36.2  & 59.4  & 65.0   & 48.6  & 56.5  & 60.2 & 52.1  & 37.8  & 67.1  & 59.0  & 41.9  & 72.0  & 54.7 \\
     \hline
     DeepJDOT \cite{DeJDOT2018}&  48.2  & 69.2  & 74.5   & 58.5  & 69.2  & 71.1 & 56.3  & 46.0  & 76.5  & 68.0  & 52.7  & 80.9  & 64.3 \\
     \hline
    CDAN~\cite{cdan1} & 50.7 & 70.6 & 76.0 & 57.6 & 70.0 & 70.0 & 57.4 & 50.9 & 77.3 & 70.9 & 56.7 & 81.6 & 65.8 \\
     \hline
     SymNets~\cite{TanMK2019} & 47.7 & 72.9 & 78.5 & 64.2 & 71.3 & 74.2 & 64.2 & 48.8 & 79.5 & 74.5 & 52.6 & 82.7 & 67.6 \\
     \hline
     ATM \cite{Atm2021} & 52.4  & 72.6  & 78.0   & 61.1  & 72.0  & 72.6 & 59.5  & 52.0  & 79.1  & 73.3  & 58.9  & 83.4  & 67.9 \\
     \hline
     DMP \cite{YouWeiPami1}&  52.3  & 73.0  & 77.3   & 64.3  & 72.0  & 71.8 & 63.6  & 52.7  & 78.5  & 72.0  & 57.7  & 81.6  & 68.1 \\
     \hline
     SCDA \cite{SCDA2021} & 57.5 & 76.9 & 80.3 & 65.7 & 74.9 & 74.5 & 65.5 & 53.6 & 79.8 & 74.5 & 59.6 & 83.7 & 70.5 \\
     \hline
     TCM \cite{TCM2021} & 58.6 & 74.4 & 79.6 & 64.5 & 74.0 & 75.1 & 64.6 & 56.2 & 80.9 & 74.6 & 60.7 & 84.7 & 70.7 \\
     \hline
     BJDA (ours)&  52.9 & 72.1 & 78.2 & 59.9 & 72.8 & 74.2 & 61.1 & 50.8 & 78.8 & 69.6 & 52.4 & 84.1 & 67.2 \\
     \hline
     BJDA+PL (ours)&  53.0 & 74.8 & 78.4 & 60.1 & 73.6 & 75.0 & 62.2 & 51.9 & 79.9 & 69.2 & 55.4 & 84.3 & 68.1 \\
    \hline
    \end{tabular}}
    \label{TabHome}
\end{table*}

\begin{table*}[htbp]
  \setlength{\abovecaptionskip}{0.cm}
  \setlength{\belowcaptionskip}{0.cm}
  \caption{Accuracy (\%) comparison between BJDA and other UDA methods on DomainNet (ResNet101). In each sub-table, the column-wise domains are selected as the source domain and the row-wise domains are selected as the target domain.}
 \centering
 \resizebox{\textwidth}{!}{
 \setlength{\tabcolsep}{0.5mm}{
   \begin{tabular}{c|ccccccc|c|ccccccc|c|ccccccc}
   \hline
   Source-only & clp   & inf   & pnt   & qdr   & rel   & skt   & Avg. & DANN~\cite{Yaroslav2016JMLR}& clp   & inf   & pnt   & qdr   & rel   & skt   & Avg.  & MCD~\cite{MCD2018} & clp   & inf   & pnt   & qdr   & rel   & skt   & Avg.  \\
   \hline
   \hline
   clp   & -     & 19.3  & 37.5  & 11.1  & 52.2  & 41.0  & 32.2  & clp   & -    & 15.5 & 34.8 & 9.5  & 50.8 & 41.4 & 30.4   & clp   & -    & 23.6 & 34.4 & 15.0 & 42.6 & 41.2 & 31.4  \\
   inf   & 30.2  & -     & 31.2  & 3.6   & 44.0  & 27.9  & 27.4  & inf   & 31.8 & -    & 30.2 & 3.8  & 44.8 & 25.7 & 27.3   & inf   & 14.2 & -    & 14.8 & 3.0  & 19.6 & 13.7 & 13.1  \\
   pnt   & 39.6  & 18.7  & -     & 4.9   & 54.5  & 36.3  & 30.8  & pnt   & 39.6 & 15.1 & -    & 5.5  & 54.6 & 35.1 & 30.0   & pnt   & 26.1 & 21.2 & -    & 7.0  & 42.6 & 27.6 & 24.9  \\
   qdr   & 7.0   & 0.9   & 1.4   & -     & 4.1   & 8.3   & 4.3   & qdr   & 11.8 & 2.0  & 4.4  & -    & 9.8  & 8.4  & 7.3    & qdr   & 1.6  & 1.5  & 1.9  & -    & 11.5 &  3.8 & 4.6   \\
   rel   & 48.4  & 22.2  & 49.4  & 6.4   & -     & 38.8  & 33.0  & rel   & 47.5 & 17.9 & 47.0 & 6.3  & -    & 37.3 & 31.2   & rel   & 45.0 & 14.2 & 45.0 & 2.2  & -    & 34.8 & 28.2  \\
   skt   & 46.9  & 15.4  & 37.0  & 10.9  & 47.0  & -     & 31.4  & skt   & 47.9 & 13.9 & 34.5 & 10.4 & 46.8 & -    & 30.7   & skt   & 33.8 & 18.0 & 28.4 & 10.2 & 29.3 & -    & 23.9  \\
   Avg.  & 34.4  & 15.3  & 31.3  & 7.4   & 40.4  & 30.5  & 26.6  & Avg.  & 35.7 & 12.9 & 30.2 & 7.1  & 41.4 & 29.6 & 26.1   & Avg.  & 26.5 & 15.4 & 26.1 & 7.4  & 31.7 & 25.7 & 20.9  \\
   \hline
   \hline
   ADDA~\cite{Tzeng2017CVPRadda}  & clp   & inf   & pnt   & qdr   & rel   & skt   & Avg.  & CDAN~\cite{cdan1}& clp   & inf   & pnt   & qdr   & rel   & skt   & Avg.  &  MarginDD~\cite{MDDlong2019} & clp   & inf   & pnt   & qdr   & rel   & skt   & Avg.  \\
   \hline
   \hline
   clp   &  -   & 11.2  & 24.1  & 3.2   & 41.9  & 30.7  & 22.2  & clp  &  -   & 20.4 & 36.6 & 9.0  & 50.7 & 42.3 & 31.8  & clp   & -    & 20.5 & 40.7 & 6.2  & 52.5 & 42.1 & 32.4 \\
   inf   & 19.1 & -     & 16.4  & 3.2   & 26.9  & 14.6  & 16.0  & inf  & 27.5 &  -   & 25.7 & 1.8  & 34.7 & 20.1 & 22.0  & inf   & 33.0 & -    & 33.8 & 2.6  & 46.2 & 24.5 & 28.0 \\
   pnt   & 31.2 & 9.5   & -     & 8.4   & 39.1  & 25.4  & 22.7  & pnt  & 42.6 & 20.0 &  -   & 2.5  & 55.6 & 38.5 & 31.8  & pnt   & 43.7 & 20.4 & -    & 2.8  & 51.2 & 41.7 & 32.0 \\
   qdr   & 15.7 & 2.6   & 5.4   & -     & 9.9   & 11.9  & 9.1   & qdr  & 21.0 & 4.5  & 8.1  &  -   & 14.3 & 15.7 & 12.7  & qdr   & 18.4 & 3.0  & 8.1  & -    & 12.9 & 11.8 & 10.8 \\
   rel   & 39.5 & 14.5  & 29.1  & 12.1   & -    & 25.7  & 24.2  & rel  & 51.9 & 23.3 & 50.4 & 5.4  &  -   & 41.4 & 34.5  & rel   & 52.8 & 21.6 & 47.8 & 4.2  & -    & 41.2 & 33.5 \\
   skt   & 35.3 & 8.9   & 25.2  & 14.9  & 37.6  & -     & 25.4  & skt  & 50.8 & 20.3 & 43.0 & 2.9  & 50.8 &  -   & 33.6  & skt   & 54.3 & 17.5 & 43.1 & 5.7  & 54.2 & -    & 35.0 \\
   Avg.  & 28.2 & 9.3   & 20.1  & 8.4   & 31.1  & 21.7  & 19.8  & Avg. & 38.8 & 17.7 & 32.8 & 4.3  & 41.2 & 31.6 & 27.7  & Avg.  & 40.4 & 16.6 & 34.7 & 4.3  & 43.4 & 32.3 & 28.6 \\
   \hline
   \hline
   \tabincell{c}{SCDA~\cite{SCDA2021}} & clp   & inf   & pnt   & qdr   & rel   & skt   & Avg.  & \tabincell{c}{BJDA\\(ours)} & clp   & inf   & pnt   & qdr   & rel   & skt   & Avg.  & \tabincell{c}{BJDA+PL\\(ours)} & clp   & inf   & pnt   & qdr   & rel   & skt   & Avg. \\
   \hline
   \hline
   clp   &  -   & 18.6 & 39.3 & 5.1  & 55.0 & 44.1 & 32.4     & clp  &  -   & 16.4 & 43.2 & 10.6 & 54.9 & 46.0 & 34.2  & clp   & -    & 17.4 & \textbf{46.0} & 12.0 & \textbf{55.9} & \textbf{46.8} & \textbf{35.6} \\
   inf   & 29.6 &   -  & 34.0 & 1.4  & 46.3 & 25.4 & 27.3     & inf  & 19.6 &  -   & 31.8 & 4.9  & 45.1 & 24.4 & 25.2  & inf   & 20.3 & -    & \textbf{34.2} & \textbf{5.8}  & 46.0 & 24.6 & 26.2 \\
   pnt   & 44.1 & 19.0 &   -  & 2.6  & 56.2 & 42.0 & 32.8     & pnt  & 34.2 & 15.9 &  -   & 6.2  & 53.9 & 27.0 & 27.4  & pnt   & 38.6 & 16.7 & -    & 6.1  & 52.5 & 27.2 & 28.2 \\
   qdr   & 30.0 & 4.9  & 15.0  &  -  & 25.4 & 19.8 & 19.0     & qdr  & 28.4 & 3.4  & 14.9 &  -   & 9.9  & 8.6  & 13.1  & qdr   & 28.4 & 3.5  & \textbf{15.1} & -    & 10.0 & 8.8  & 13.2 \\
   rel   & 54.0 & 22.5 & 51.9 & 2.3  &   -  & 42.5 & 34.6     & rel  & 55.5 & 14.7 & 52.7 & 5.2  &  -   & 24.7 & 30.5  & rel   & \textbf{55.8} & 14.5 & \textbf{54.2} & 5.7  & -    & 24.5 & 30.9 \\
   skt   & 55.6 & 18.5 & 44.7 & 6.4  & 53.2 &  -   & 35.7     & skt  & 53.4 & 11.6 & 47.7 & 10.3 & 39.6 &  -   & 32.5  & skt   & \textbf{56.1} & 13.7 & \textbf{48.2} & 11.6 & 46.0 & -    & 35.1 \\
   Avg.  & 42.6 & 16.7 & 37.0 & 3.6  & 47.2 & 34.8 & 30.3     & Avg. & 38.2 & 12.4 & 38.1 & 7.4  & 40.7 & 26.1 & 27.2  & Avg.  & 39.8 & 13.2 & \textbf{39.5} & 8.2  & 42.1 & 26.4 & 28.2 \\
   \hline
   \end{tabular}
   }}
 \label{TbDnet1}
 \vspace{-1mm}
\end{table*}

To evaluate the scalability of the proposed BJDA method, We compare BJDA with several recent UDA methods on the large-scale datasets Office-Home and DomainNet. During the training process, we assign pseudo-labels for the unlabeled target samples. Those pseudo-labels may be wrong, so these noises in pseudo-labels will degrade the performance of our BJDA method. To address this issue, we select the target samples which have prediction probability greater than a confidence threshold in the training stage. We add the experiments with this selection strategy on the Office-Home and DomainNet datasets. The results are shown in Table~\ref{TabHome} and Table~\ref{TbDnet1}, and the corresponding setting is denoted as BJDA+PL. It can be seen that the selection strategy for pseudo-labels can improve the performance of the proposed BJDA method and alleviate the impact of noisy pseudo-labels. For the Office-Home dataset, our BJDA outperforms the recently proposed ATM method~\cite{Atm2021} by 0.2\% and the average accuracy is close to the state-of-the-arts (i.e., SCDA~ \cite{SCDA2021} and TCM~\cite{TCM2021}). The DomainNet dataset is a large and very challenging  benchmark, in which large domain gaps exist. Our BJDA obtains an average accuracy of 28.2\%, which is very close to the best result 30.3\% produced by SCDA~\cite{SCDA2021}. The reason may be that SCDA uses multiple complicated pair-wise adversarial processes, in which the classifier first finds the region with a large discrepancy among samples with the same category, and then the feature extractor tries to suppress the dissimilar regions. TCM combines the disentangled causal mechanisms with deep neural networks and it can effectively identify the latent semantic feature and improve the discriminative ability of representations. In addition, our BJDA achieves the best accuracy on 12 out of the 30 UDA tasks on the DomainNet dataset. From these results, we find that BJDA achieves comparable accuracy compared with recent UDA methods on those large-scale benchmarks. These results verify the scalability and the effectiveness of the proposed BJDA method.

\subsection{Feature visualization for BJDA }

\begin{figure*}[tb]
\centering{
\subfigure[Source-only($I\rightarrow P$)]{\label{Fig.sub.1-1}
\begin{minipage}[b]{0.23\textwidth} 
\centering \scalebox{0.33}{ 
\includegraphics{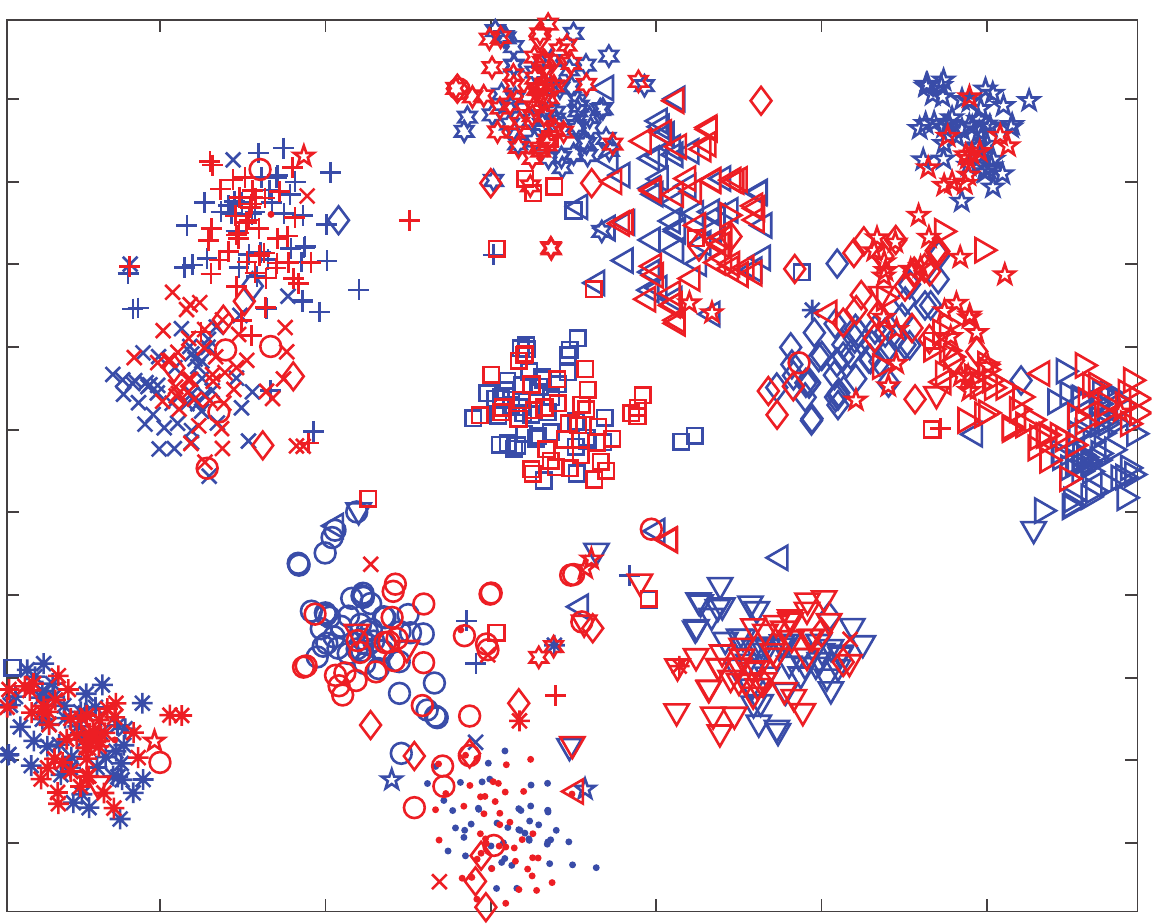}}
\end{minipage}}
\subfigure[DANN($I\rightarrow P$)]{\label{Fig.sub.1-2}
\begin{minipage}[b]{0.23\textwidth}
\centering \scalebox{0.33}{ 
\includegraphics{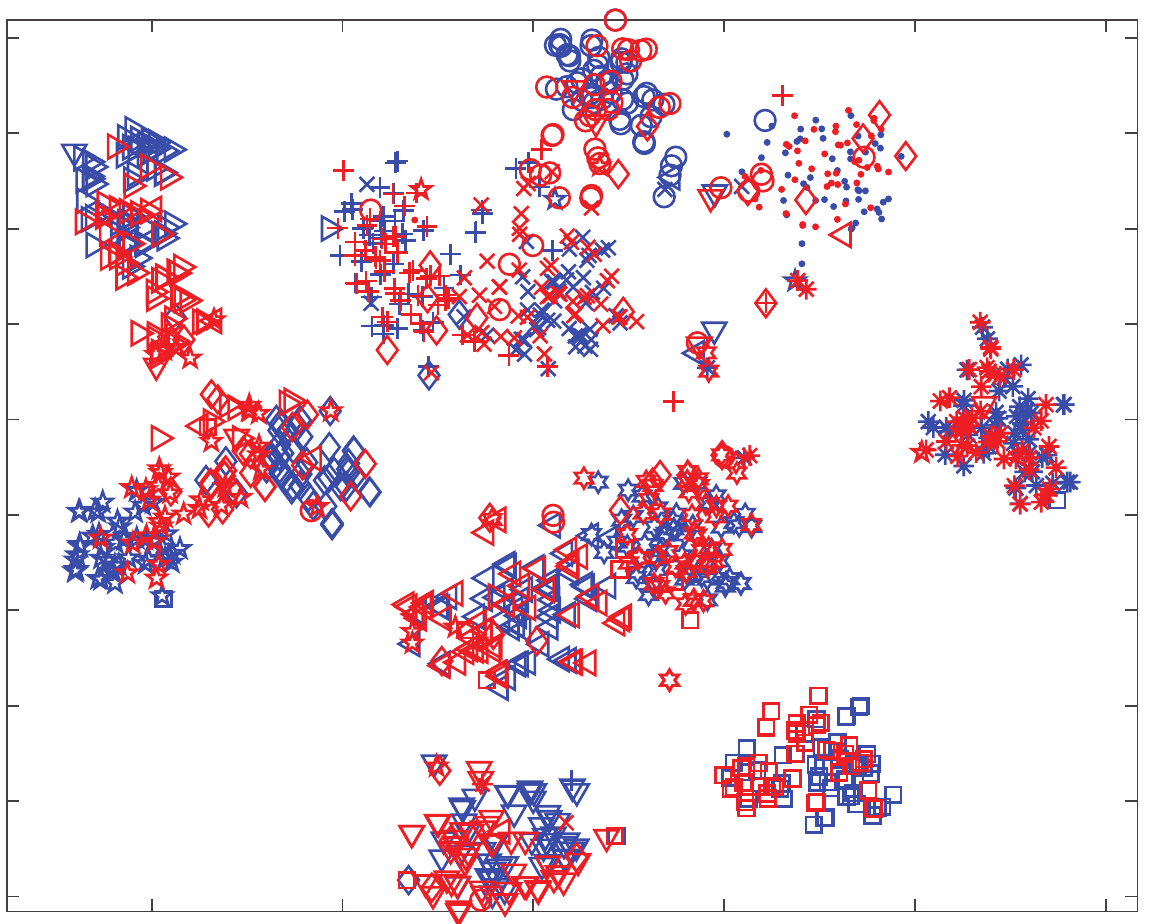}}
\end{minipage}}
\subfigure[DeepJDOT($I\rightarrow P$)]{\label{Fig.sub.1-3}
\begin{minipage}[b]{0.23\textwidth}
\centering \scalebox{0.33}{ 
\includegraphics{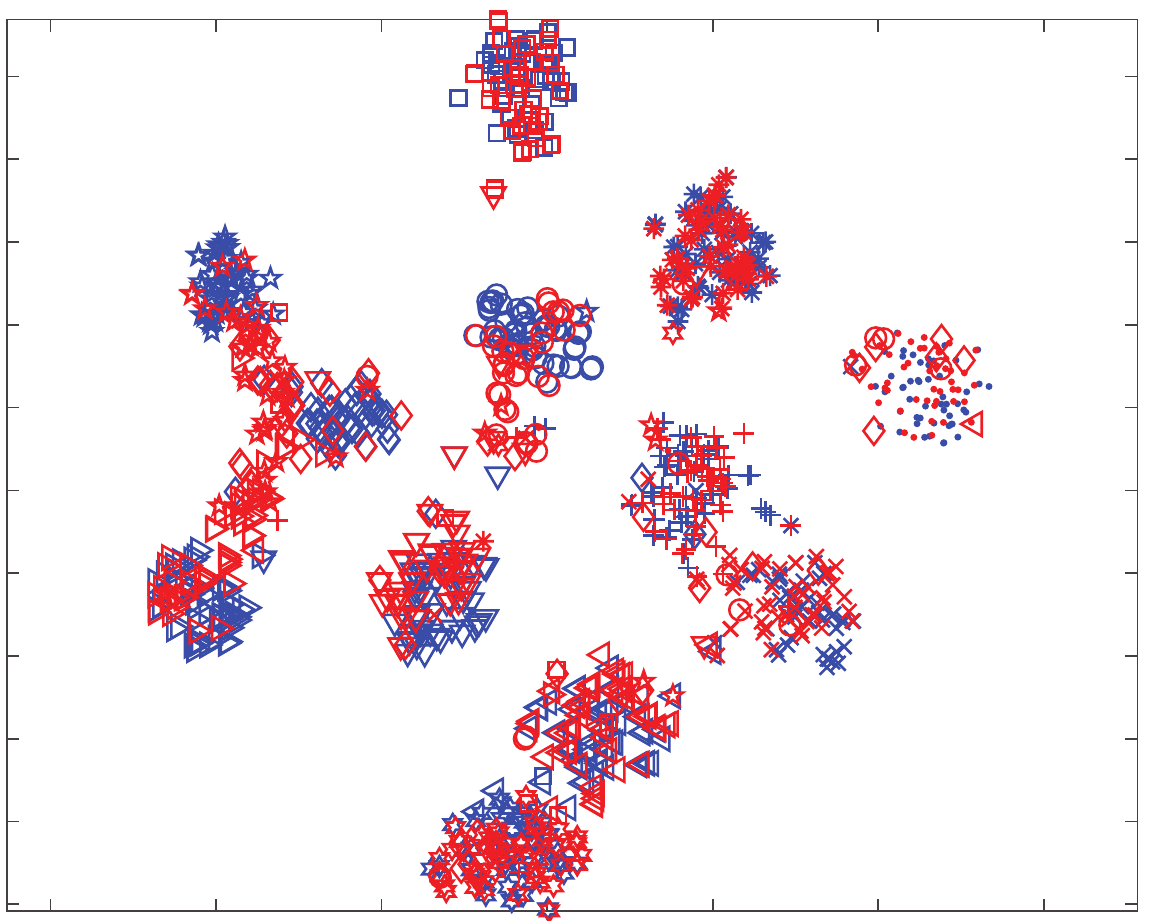}}
\end{minipage}}
\subfigure[BJDA($I\rightarrow P$)]{\label{Fig.sub.1-4}
\begin{minipage}[b]{0.23\textwidth}
\centering \scalebox{0.33}{ 
\includegraphics{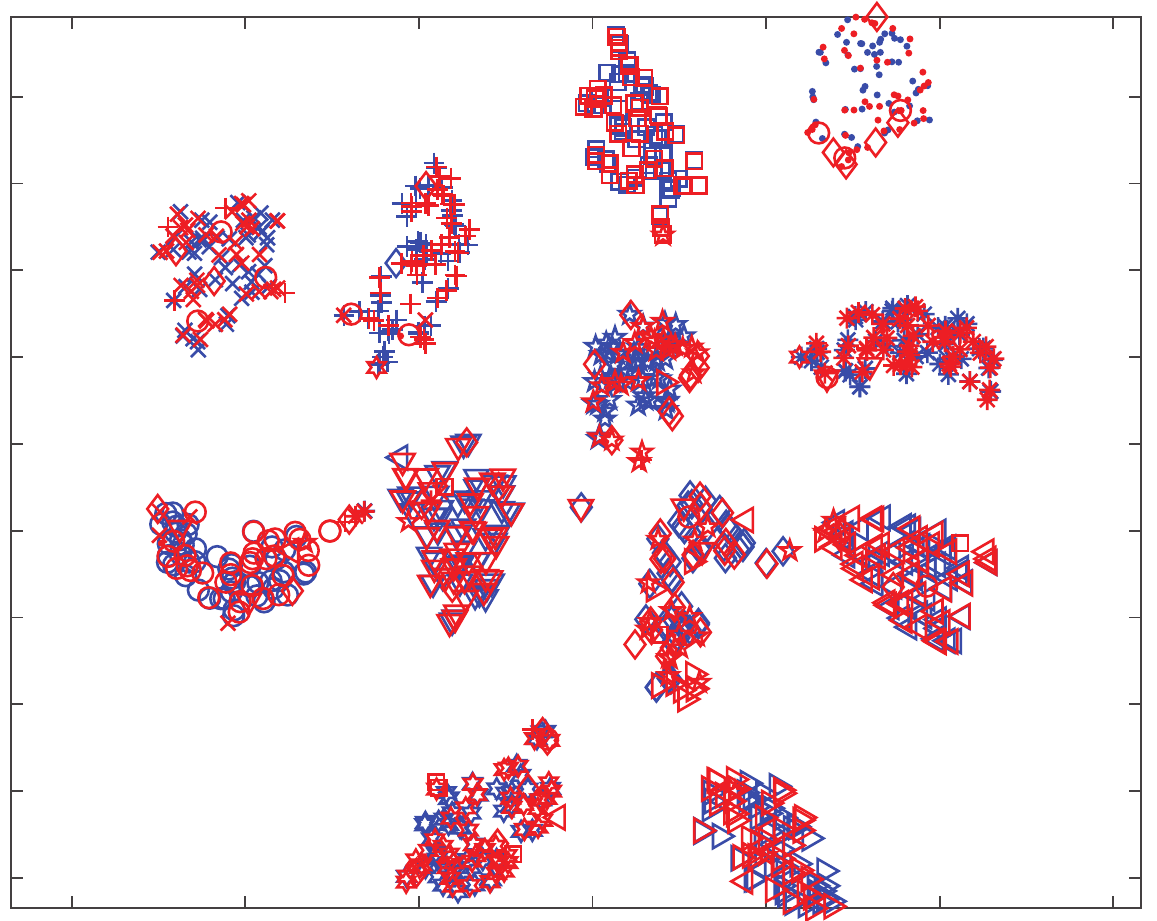}}
\end{minipage}}\\
\subfigure[Source-only($P\rightarrow I$)]{\label{Fig.sub.2-1}
\begin{minipage}[b]{0.23\textwidth}
\centering \scalebox{0.33}{ 
\includegraphics{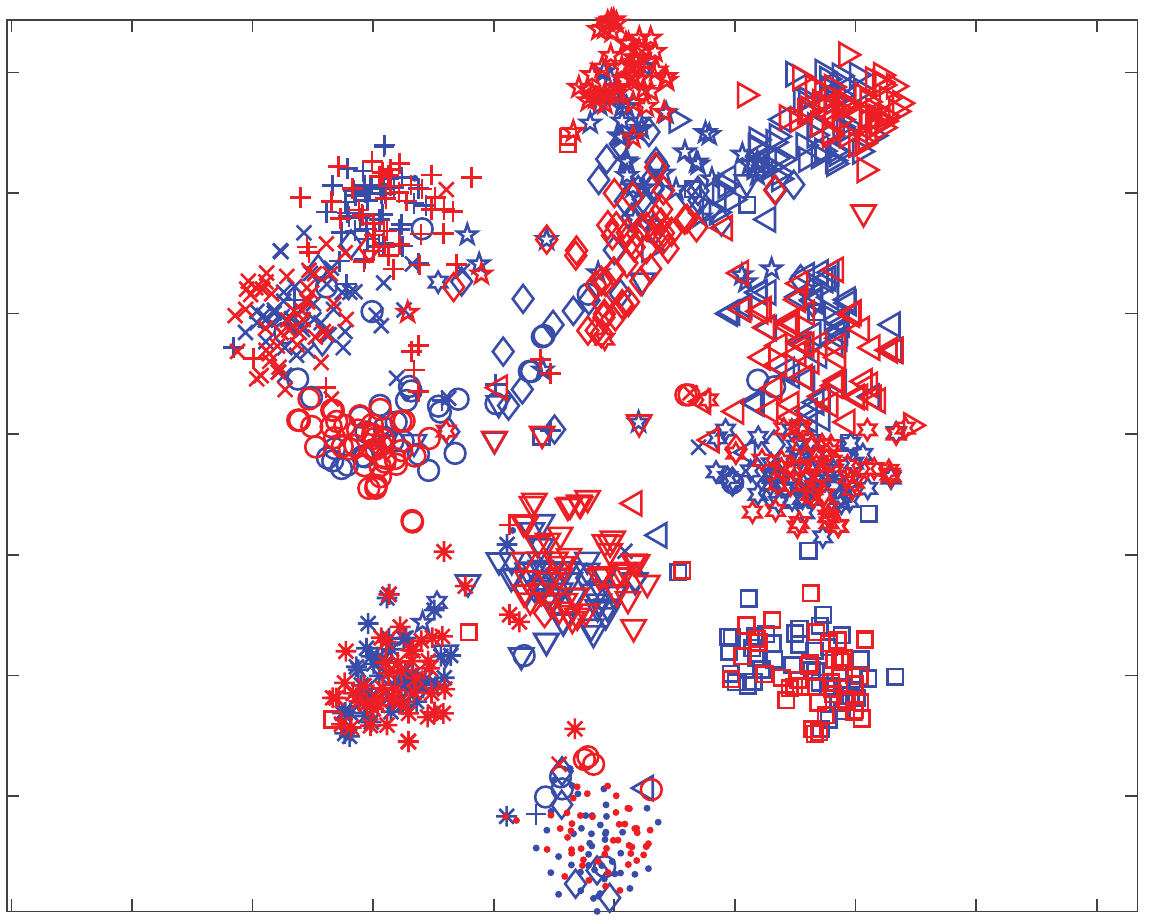}}
\end{minipage}}
\subfigure[DANN($P\rightarrow I$)]{\label{Fig.sub.2-2}
\begin{minipage}[b]{0.23\textwidth}
\centering \scalebox{0.33}{ 
\includegraphics{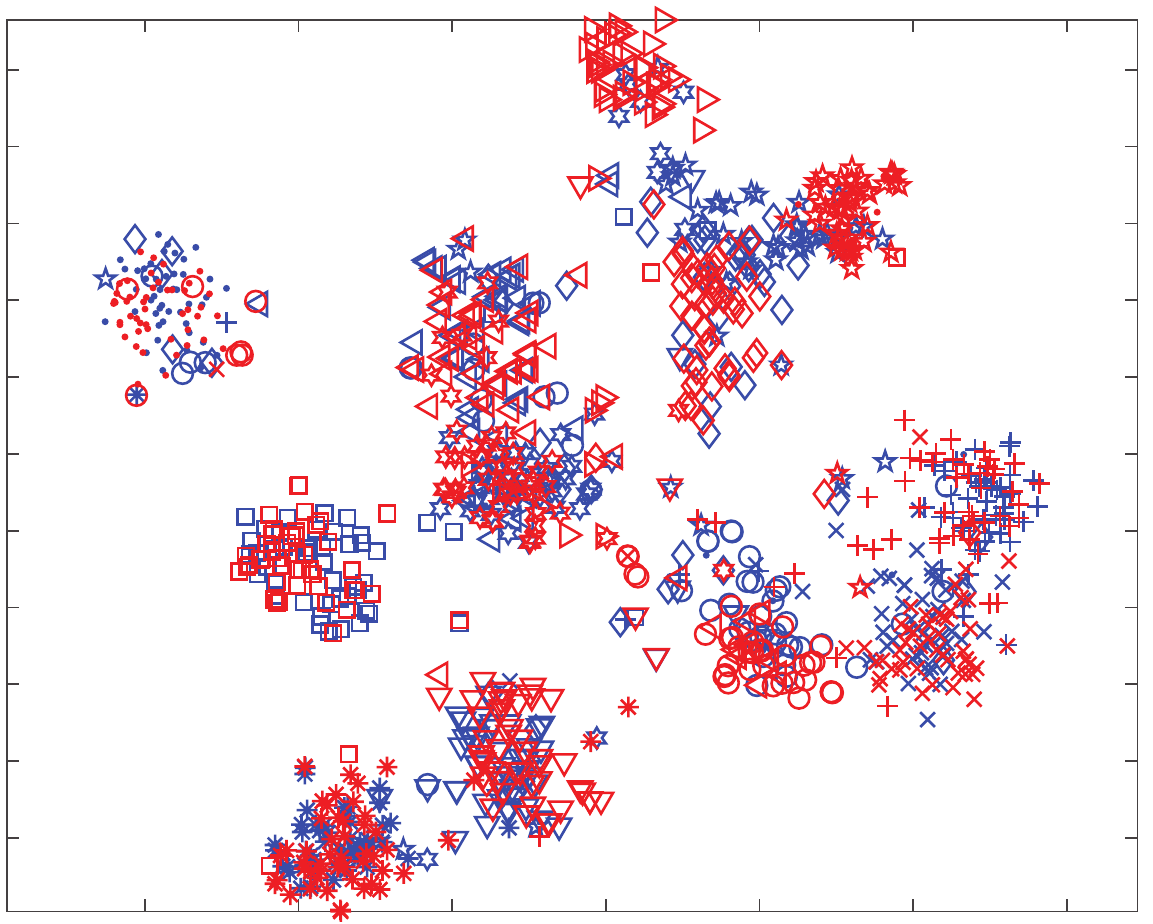}}
\end{minipage}}
\subfigure[DeepJDOT($P\rightarrow I$)]{\label{Fig.sub.2-3}
\begin{minipage}[b]{0.23\textwidth}
\centering \scalebox{0.33}{ 
\includegraphics{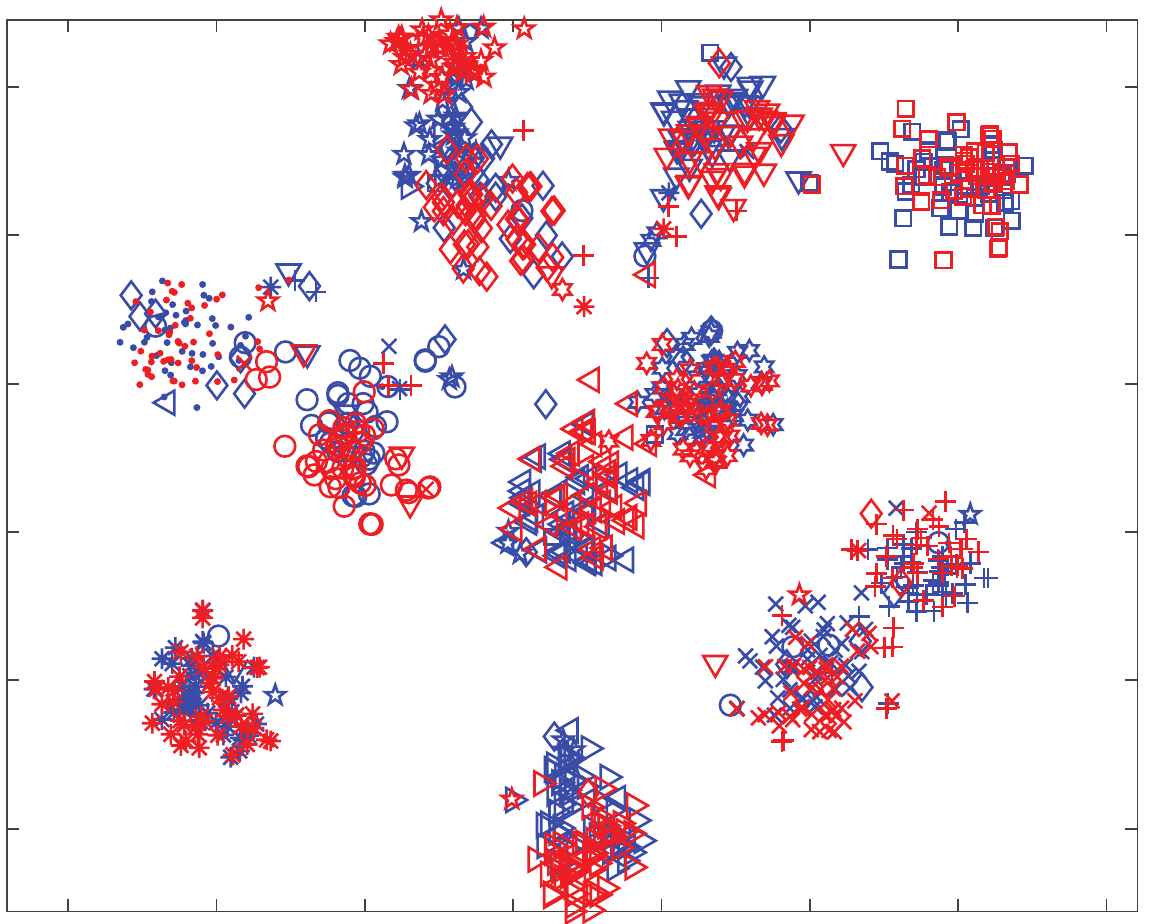}}
\end{minipage}}
\subfigure[BJDA($P\rightarrow I$)]{\label{Fig.sub.2-4}
\begin{minipage}[b]{0.23\textwidth}
\centering \scalebox{0.33}{ 
\includegraphics{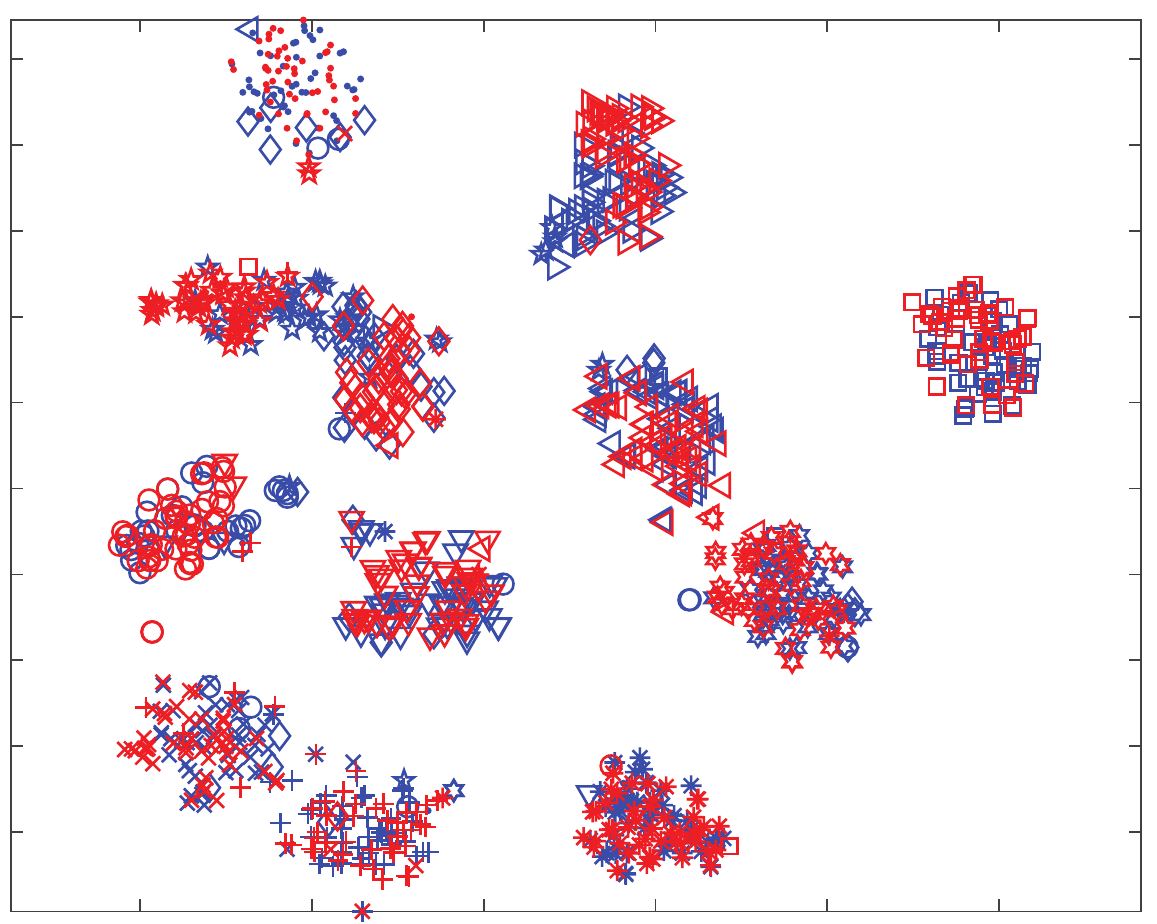}}
\end{minipage}}
\caption{Visualization of the learned features by t-SNE\cite{Maaten2008Visualizing}. The tasks I$\rightarrow$P and P$\rightarrow$I in ImageCLEF-DA are used as examples. Specifically, the Source-only features (non-adapted), the DANN features, the DeepJDOT features and our BJDA features are visualized in each column, respectively. The source features is shown in blue and target in red. We use different markers to denote different categories. Better viewed in color.}\label{tsne1}}
\end{figure*}

To show the effectiveness of BJDA intuitively, we visualize the deep features before and after adaptation using t-SNE \cite{Maaten2008Visualizing} in Fig. \ref{tsne1}. The experiment of tasks I$\rightarrow$P and P$\rightarrow$I in ImageCLEF-DA are selected for illustration. Features extracted from DANN \cite{Yaroslav2016JMLR} and DeepJDOT \cite{DeJDOT2018} are also shown in Fig. \ref{tsne1} for comparison. For clarity, we use different markers to denote different categories, and the source features are shown in blue and target in red. Compared with the features learned by the Source-only in Fig. \ref{Fig.sub.1-1} and \ref{Fig.sub.2-1}, we observe that all the features from BJDA in Fig. \ref{Fig.sub.1-4} and \ref{Fig.sub.2-4} have shown good adaptation patterns, which implies that BJDA is helpful to reduce the domain shift and improve the transferability of deep features. We can see that the features of DeepJDOT \cite{DeJDOT2018} in the third column are more separable than those of DANN \cite{Yaroslav2016JMLR} in the second column, which demonstrates that it is more valuable to align the joint distributions compared with the marginal-based methods. As shown in Fig. \ref{tsne1}, there are larger margins between different clusters corresponding to different categories in Fig. \ref{Fig.sub.1-4} and \ref{Fig.sub.2-4} compared with graphs in other columns, and the target features in the fourth column behave more clear to categorize, which indicates that BJDA can learn more discriminative features. It is shown that the features of BJDA are better aligned and more distinguishable than the features of DeepJDOT \cite{DeJDOT2018}, which indicates the superiority of performing joint distribution alignment in RKHS.

\subsection{Ablation Study}
\begin{table*}[htbp]
  \centering
  \caption{Ablation study of the proposed BJDA on ImageCLEF-DA (ResNet50).}
    \begin{tabular}{cccccccc}
    \hline
    Method & I$\rightarrow$P  & P$\rightarrow$I  & I$\rightarrow$C  & C$\rightarrow$I  & C$\rightarrow$P  & P$\rightarrow$C  & Avg \\
    \hline
    Source-only &74.8$\pm$0.3  & 83.9$\pm$0.1  & 91.5$\pm$0.3  & 78.0$\pm$0.2  & 65.5$\pm$0.3  & 91.2$\pm$0.3  & 80.7 \\
    \hline
    DeepJDOT \cite{DeJDOT2018} &  79.4&86.7&93.2&82.0&69.5&91.3&83.7 \\
    \hline
    TripDA \cite{TripDA} &  78.1$\pm$0.3  & 89.2$\pm$0.1 & 96.8$\pm$0.2  & 91.3$\pm$0.2  & 78.2$\pm$0.2  & 94.0$\pm$0.3  & 87.9  \\
    \hline
    BJDA($w/o$ $L_{da}$) & 80.8$\pm$0.5  & 91.5$\pm$0.2  & 95.1$\pm$0.3  & 92.3$\pm$0.3  & 77.4$\pm$0.2  & 95.3$\pm$0.2  & 88.7 \\
    \hline
    BJDA($w/o$ $L_{dmc}$) & 81.7$\pm$0.2  & 93.3$\pm$0.5  & 97.0$\pm$0.4  & 93.0$\pm$0.4  & 81.5$\pm$0.3  & 95.5$\pm$0.5  & 90.3\\
    \hline
    BJDA($w/o$ $L_{dmc}$)+$L_{trip}$ & 78.5$\pm$0.6  & 88.3$\pm$0.4  & 96.3$\pm$0.3  & 95.2$\pm$0.3  & 80.4$\pm$0.5  & 95.8$\pm$0.2  & 89.1\\
    \hline
    BJDA($w/o$ $L_{da}$)+$WD$ & 75.7$\pm$0.1  & 84.6$\pm$0.0  & 92.9$\pm$0.0  & 91.6$\pm$0.0  & 77.5$\pm$0.1  & 92.5$\pm$0.0  & 85.8\\
    \hline
    BJDA($w/o$ $L_{da}$)+$MDD$ & 80.0$\pm$0.0  & 92.0$\pm$0.1  & 96.0$\pm$0.0  & 92.6$\pm$0.0  & 78.7$\pm$0.0  & 95.4$\pm$0.1  & 89.1\\
    \hline
    BJDA($w/o$ $L_{da}$)+$LMMD$ & 80.3$\pm$0.1  & 91.0$\pm$0.1  & 95.6$\pm$0.0  & 92.3$\pm$0.0  & 78.1$\pm$0.2  & 93.9$\pm$0.1  & 88.5\\
    \hline
    BJDA & \textbf{81.9$\pm$0.4}  & \textbf{93.8$\pm$0.1}  & \textbf{97.5$\pm$0.4}  & \textbf{95.8$\pm$0.2}  & \textbf{81.9$\pm$0.1}  & \textbf{96.3$\pm$0.1}  & \textbf{91.2} \\
    \hline
    \end{tabular}%
  \label{Abla}%
\end{table*}
In order to exemplify the performance of BJDA and better investigate where the improvements are coming from, we perform an ablation study over various combinations for the UDA task on ImageCLEF-DA. The BJDA algorithm reduces the differences in the joint distributions and learns a category discriminative feature space simultaneously. It mainly consists of two components: the Bures joint distribution alignment loss and the dynamic margin contrastive loss. The BJDA($w/o$ $L_{dmc}$) implements the idea of Bures joint distribution alignment but excludes the DMC loss. The BJDA($w/o$ $L_{da}$) trains the model with the cross-entropy loss and the DMC loss. The BJDA implements the proposed BJDA algorithm as Eq. (\ref{eq8}). DeepJDOT \cite{DeJDOT2018} incorporates OT and deep neural networks in the Euclidean space to matches the joint distributions, we select DeepJDOT as the baseline to demonstrate the superiority of our Bures joint distribution alignment loss. In light of the original triplet loss \cite{Trip2015} is widely used to preserve the semantic relations, we train our network using the Bures joint distribution alignment loss, the cross-entropy loss, and the triplet loss, and is termed as BJDA($w/o$ $L_{dmc}$)+$L_{trip}$. We select BJDA($w/o$ $L_{dmc}$)+$L_{trip}$ as the baseline to show the effectiveness of our DMC loss. The recently proposed TripDA \cite{TripDA} uses the original triplet loss to model the semantic structure of the data and incorporates the Joint Maximum Mean Discrepancy (JMMD) \cite{Long2017JAN} to reduce the domain shift. We compare BJDA($w/o$ $L_{dmc}$)+$L_{trip}$ with TripDA \cite{TripDA} to show the effectiveness of our Bures joint distribution alignment loss. Table \ref{Abla} shows the results of the ablation study.

Comparing BJDA($w/o$ $L_{dmc}$) with DeepJDOT \cite{DeJDOT2018}, it is clear that on average the BJDA($w/o$ $L_{dmc}$) performs better, which demonstrates the superiority of our Bures joint distribution alignment loss that performs domain alignment via OT in RKHS. BJDA($w/o$ $L_{dmc}$)+$L_{trip}$ outperforms the TripDA \cite{TripDA} by a clear margin, which verifies the effectiveness of the proposed Bures joint distribution alignment loss compared with JMMD. Our BJDA achieves 2.1\% improvement on the average accuracy against BJDA($w/o$ $L_{dmc}$)+$L_{trip}$. This result shows the superiority of the DMC loss compared with the original triplet loss \cite{Trip2015} and indicates that our DMC loss can be of help to improve the category discriminative ability of feature space. Notice that BJDA($w/o$ $L_{dmc}$) obtains an improvement of 1.6\% over BJDA($w/o$ $L_{da}$), which reveals that the Bures joint distribution alignment loss makes a greater contribution to the UDA task compared with the DMC loss. From the above results, we can confirm that each component in BJDA has a specific contribution. Combining the Bures joint distribution alignment loss and the DMC loss leads to further improvement, indicating their complementarity and superiority of the BJDA algorithm.

LMMD~\cite{DSAN2020} is a recently proposed divergence for conditional MMD, which models the domain divergence by aligning the marginal distributions of subdomains. We compare our BJDA with conditional MMD by substituting the Eq. (\ref{eq5}) loss with the LMMD loss. The setting is denoted as BJDA($w/o$ $L_{da}$)+$LMMD$. It can be seen that the Eq. (\ref{eq5}) loss performs better than LMMD, which verifies the superiority of aligning joint distributions. MDD~\cite{Atm2021} is another domain divergence, which considers the marginal distributions by minimizing the inter-domain divergence and optimizes the conditional distributions by introducing class information. We empirically evaluate the proposed BJDA by substituting the Eq. (\ref{eq5}) loss with the MDD loss. The setting is denoted as BJDA($w/o$ $L_{da}$)+$MDD$. It can be seen that the Eq. (\ref{eq5}) loss achieves better results compared with the MDD loss. We also add the ablation experiment with the variant which uses the Wasserstein distance on the ImageCLEF-DA dataset. We compute the Wasserstein distance follow the formula of discrete Optimal Transport from previous work~\cite{OTrkhs}. The corresponding setting is denoted as BJDA($w/o$ $L_{da}$)+$WD$. The results suggest that the BJDA loss gets better performance than the Wasserstein distance. These results further verify the superiority of the proposed Bures joint distribution alignment loss in Eq. (\ref{eq5}).
\subsection{Convergence Analysis}
\begin{figure}[t]
\subfigure[Classification loss of I$\rightarrow$P]{\label{line1.sub.m1}
\begin{minipage}[b]{0.23\textwidth} 
\centering \scalebox{0.225}{ 
\includegraphics{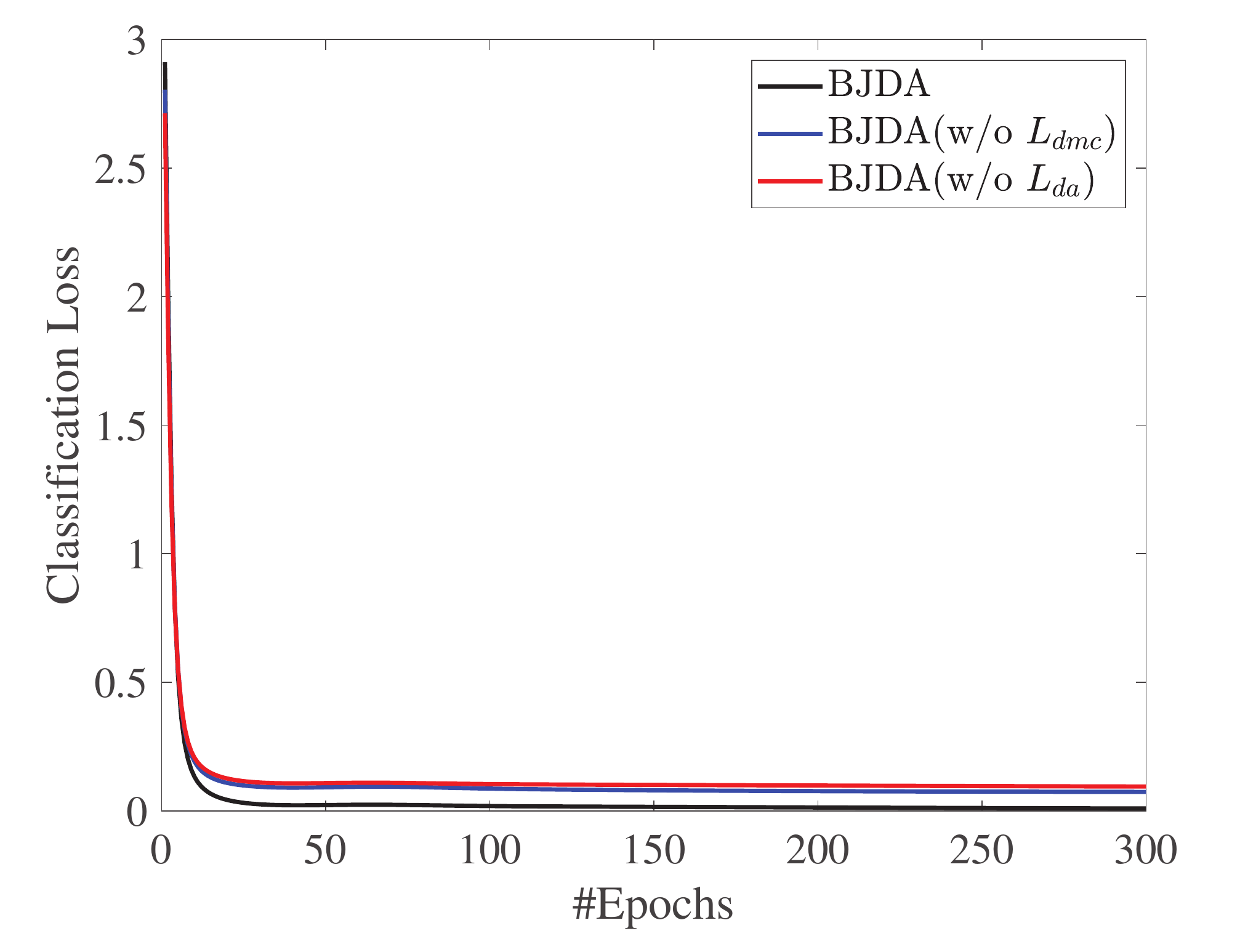}}
\end{minipage}}
\subfigure[Total loss of I$\rightarrow$P]{\label{line1.sub.m2}
\begin{minipage}[b]{0.23\textwidth}
\centering \scalebox{0.225}{ 
\includegraphics{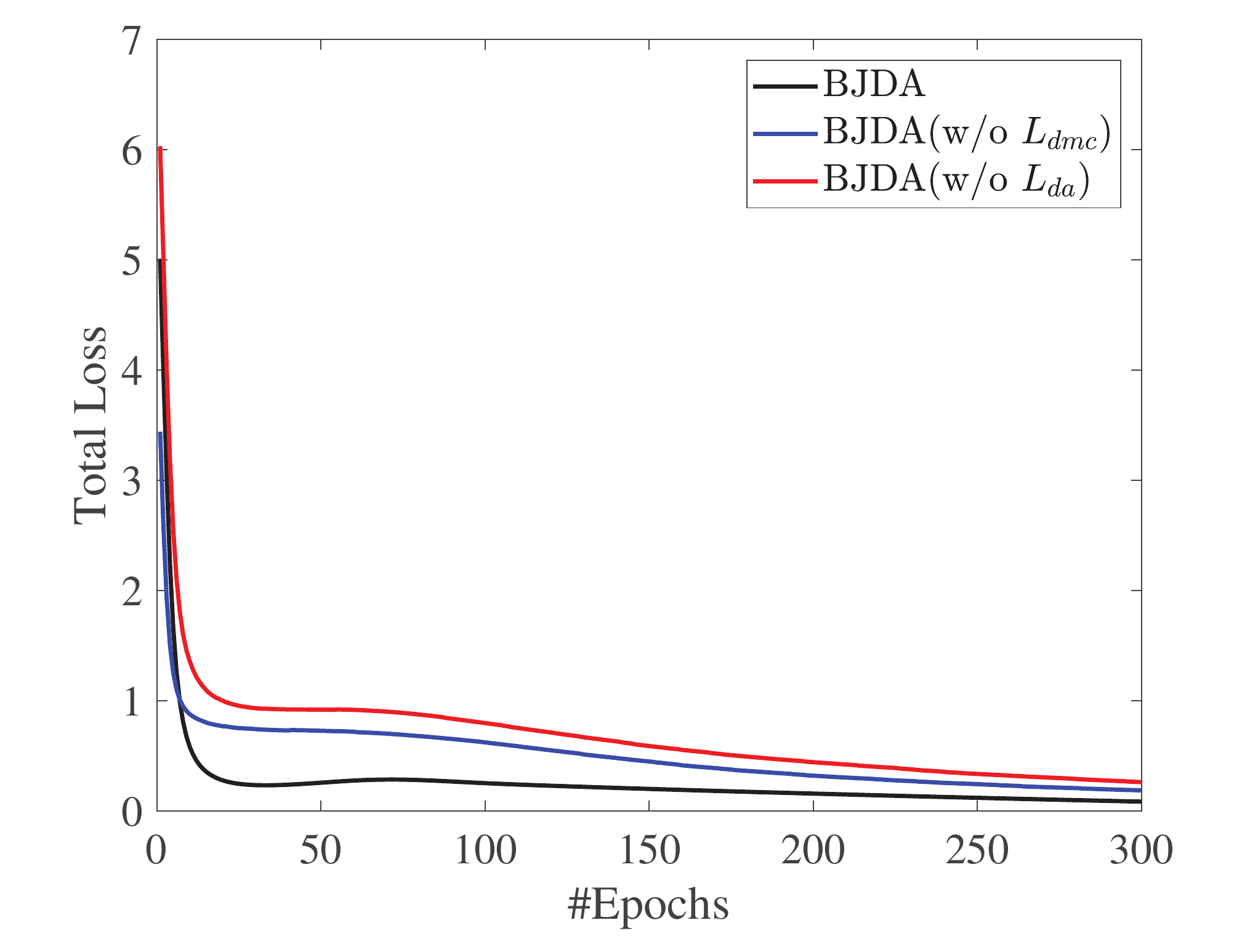}}
\end{minipage}}
\caption{The classification loss (left) and the total loss (right) comparisons between BJDA, BJDA(w/o $L_{dmc}$), BJDA(w/o $L_{da}$). The evaluation of the task I$\rightarrow$P in ImageCLEF-DA is used as an example. Better viewed in color.}\label{line1}
\end{figure}
To show the stability of training, we report the classification loss values and total loss values during the epochs in Fig. \ref{line1}. To reveal the effectiveness of our contribution, we compare the loss values between BJDA, BJDA($w/o$ $L_{dmc}$), and BJDA($w/o$ $L_{da}$). We take the task I$\rightarrow$P in ImageCLEF-DA as an example. It can be seen that BJDA gradually converges after 300 epochs. Fig. \ref{line1.sub.m1} show that the classification loss values of BJDA (black line) are lower than that of BJDA($w/o$ $L_{dmc}$) (blue line) and BJDA($w/o$ $L_{da}$) (red line), which indicates that the features learned by BJDA are more discriminative. Fig. \ref{line1.sub.m2} presents the comparison of total loss values, it can be seen that the values of the total loss of BJDA are much smaller, which demonstrates the superiority of BJDA, and BJDA is better towards the convergence.

\subsection{Hyper-parameters Analysis}
\begin{figure}[t]
\subfigure[Hyper-parameters of I$\rightarrow$P]{\label{Hypm1.sub.m1}
\begin{minipage}[b]{0.23\textwidth} 
\centering \scalebox{0.4}{ 
\includegraphics{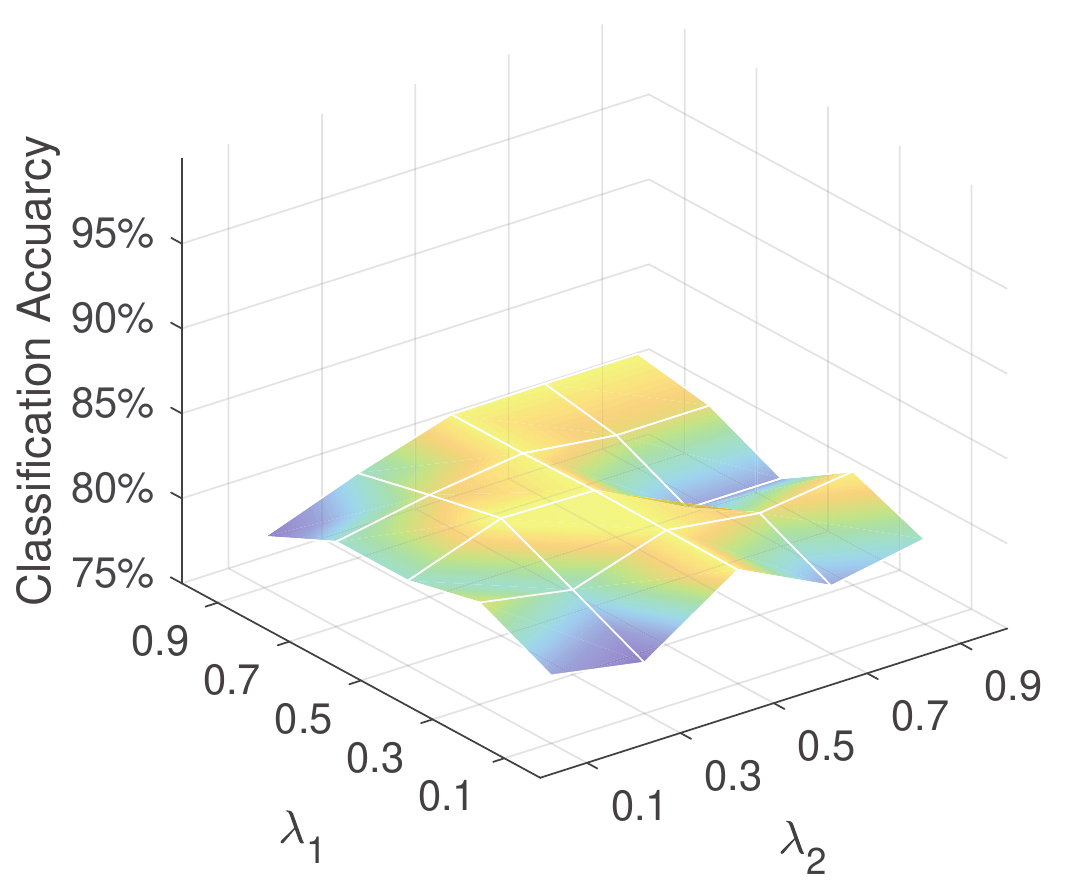}}
\end{minipage}}
\subfigure[Hyper-parameters of P$\rightarrow$I]{\label{Hypm1.sub.m2}
\begin{minipage}[b]{0.23\textwidth}
\centering \scalebox{0.4}{ 
\includegraphics{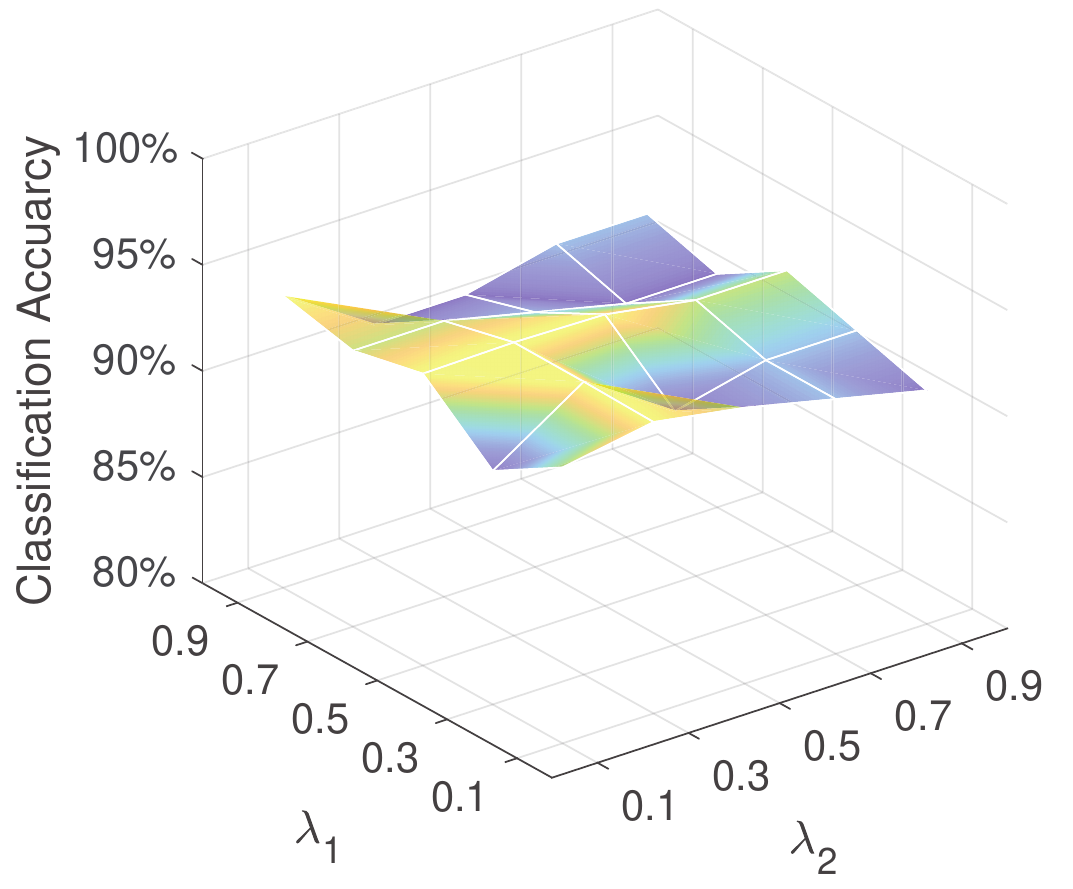}}
\end{minipage}}
\caption{The impact of hyper-parameters $\lambda_1$ and $\lambda_2$, the tasks I$\rightarrow$P and P$\rightarrow$I in ImageCLEF-DA are used as examples. Better viewed in color.}\label{Hypm1}
\end{figure}
The overall objective function, i.e., Eq. (\ref{eq8}), of the proposed BJDA includes two hyper-parameters $\lambda_1$ and $\lambda_2$ to balance the Bures joint distribution alignment loss and the dynamic margin contrastive loss. To show the stability of our BJDA methods and the impact of different hyper-parameters $\lambda_1$ and $\lambda_2$, we investigate the sensitivity of the hyper-parameters and implement grid search for $\lambda_1$ and $\lambda_2$ with respect to the tasks I$\rightarrow$P and P$\rightarrow$I in ImageCLEF-DA. $\lambda_1$ and $\lambda_2$ are searched in the range of [0.1, 0.3, 0.5, 0.7, 0.9], respectively. Results are shown in Fig. \ref{Hypm1}. The brighter the color, the higher the classification accuracy is. It can be seen that the classification accuracy varies slowly among the peak area. The results show that in most cases, the performance of BJDA is stable when the values of $\lambda_1$ and $\lambda_2$ increase from 0.1 to 0.9. This indicates the robustness of performing joint distribution alignment in the RKHS. The values of $(\lambda_1, \lambda_2)$ in this paper are set as (0.5,0.3). The fact that $\lambda_1 > \lambda_2$ indicates that the Bures joint distribution alignment phase is more important than the metric learning phase during the adaptation process.

\section{Conclusion}\label{section5}
In this paper, we propose a novel method called Bures joint Distribution Alignment (BJDA) for unsupervised domain adaptation. We directly align the joint distributions of the two domains based on the optimal transport theory. To empower the optimal transport theory with the capability of capturing nonlinear structures underlying the data, our work minimizes the kernel Bures-Wasserstein distance of the joint distributions in the reproducing kernel Hilbert space. In addition, we introduce a dynamic margin based contrastive learning phase to improve the category discriminative ability of representations. Extensive experiments on four benchmarks verify that BJDA exceeds previous state-of-the-art methods with significant advantages. While BJDA is designed for UDA task with a single source domain, there exist some works among multiple source domains. In the future, we plan to extend BJDA to the scenario with multiple source domains.
\bibliographystyle{IEEEtran}
\bibliography{BJDA_ie3}

\end{document}